# The Michigan Robotics Undergraduate Curriculum

*Defining the Discipline of Robotics for Equity and Excellence*

**Odest Chadwicke Jenkins**
Professor of Robotics
and Electrical Engineering & Computer Science

Associate Director for Undergraduate Education,
Robotics Institute

**Jessy Grizzle**
Professor of Robotics

Director, Robotics Institute (2016-22)

**Ella Atkins**
Associate Director for Graduate Programs,
Robotics Institute (2013-20)

**Leia Stirling**
Associate Professor of Robotics
and Industrial Engineering

**Elliott Rouse**
Associate Professor of Robotics
and Mechanical Engineering

**Mark Guzdial**
Professor of Electrical Engineering & Computer Science

Professor, Engineering Education Research Program

**Damen Provost**
Managing Director, Robotics Institute

**Kimberly Mann**
Unit Administrator, Robotics Institute

**Joanna Millunchick**
Associate Dean for Undergraduate Education,
College of Engineering (2017-22)

University of Michigan

August 10, 2023

# Prologue

The Robotics Major at the University of Michigan was successfully launched in 2022 – just a year ago – as an innovative step forward to better serve students, our communities, and our society. Building on our guiding principle of "Robotics with Respect" and our larger Robotics Pathways model, the Michigan Robotics Major was designed to define robotics as a true academic discipline with both equity and excellence as our highest priorities. Understanding that talent is equally distributed but opportunity is not, the Michigan Robotics Major has embraced an adaptable curriculum that is accessible through a diversity of student pathways and enables successful and sustained career-long participation in robotics, AI, and automation professions. The results after our planning efforts (2019-22) and first academic year (2022-23) have been highly encouraging:

- More than 100 students declared Robotics as their major
- Completion of the Robotics major by our first two graduates
- Soaring enrollments in our Robotics classes
- Thriving partnerships with Historically Black Colleges and Universities

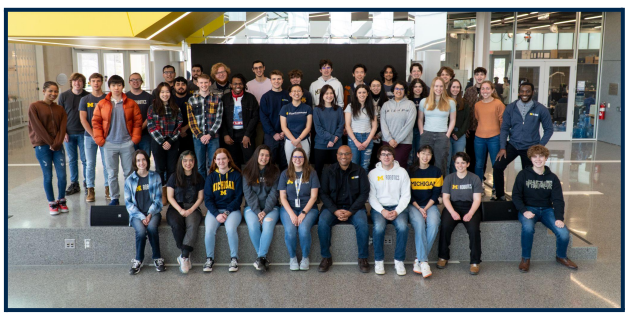
Inaugural Robotics Major Class Photo (April 2022)

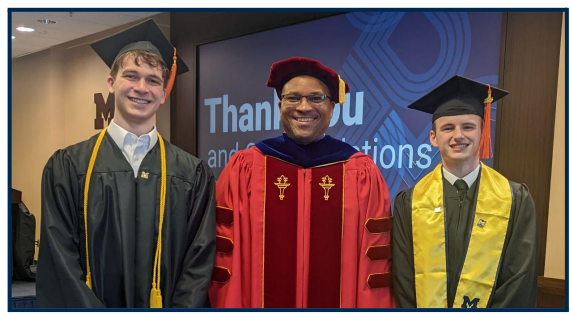
First Robotics Major graduates (April 2022): Joseph Taylor and Connor Williams with Prof. Jenkins

This document provides our original curricular proposal for the Robotics Undergraduate Program at the University of Michigan, submitted to the Michigan Association of State Universities in April 2022 and approved in June 2022. The dissemination of our program design is in the spirit of continued growth for higher education towards realizing equity and excellence.

# Acknowledgments

The Michigan Robotics Undergraduate Program owes a tremendous debt of gratitude to many people across our Robotics Institute and Robotics Department, the University of Michigan, the College of Engineering, the State of Michigan, and the greater national and global robotics community. Creating a first-of-a-kind robotics program is an incredibly bold and daring undertaking that would not be possible without the support, contributions, empathy, and insights from all corners of our amazing university (Go Blue!). While it would be impossible to recognize everyone who played important roles in realizing the Robotics Major, we would like to acknowledge some individuals who were especially critical to the formation of the program.

We must first thank Dean Alec Gallimore and the College of Engineering for their visionary leadership throughout our evolution. Under the guidance and stewardship of Dean Gallimore, the Robotics Institute was able to grow, thrive, and prove it has the right stuff to become a viable academic department and undergraduate program. None of this would be possible without your confidence in us and willingness to innovate for the Common Good.

The Robotics Institute owes its origins to Dawn Tilbury – the founding Director of the Robotics Institute (in 2014 under Dean David Munson) and now the inaugural Chair of the Robotics Department – and her foresight to envision what has become the home of Michigan Robotics – the [Ford Motor Company Robotics Building](#).

Nadine Sarter, Associate Dean Michael Wellman, and the Robotics Future Committee did tremendous work between 2018-20 to explore the potential and opportunities for Michigan to establish a department and undergraduate program in robotics. Their work identified the path for Michigan to earn distinguished leadership in robotics. The Robotics Futures Committee also underscored the deep and interdisciplinary history Michigan has in robotics – highlighting that Michigan Robotics is at its best when we bring together people and robots.

Brian Denton, Ellen Arruda, and Jen Piper provided valuable insights and perspectives through our work together during 2020-21 on the Robotics Planning Committee, led by Associate Dean Joanna Millunchick. Much of the Robotics Undergraduate Program took shape through this committee to answer the core question of "[Why do we need a Robotics Department](#)?" The authors had many stimulating and spirited discussions through this committee. These discussions, and considerable and prescient market research by Michigan Engineering Nexus, yield the essential foundation for a compelling new Robotics Major. This planning also greatly benefited from the recommendation of the joint Aerospace-Robotics Curriculum Planning Committee, led by Dimitra Panagou and created by Anthony Waas.

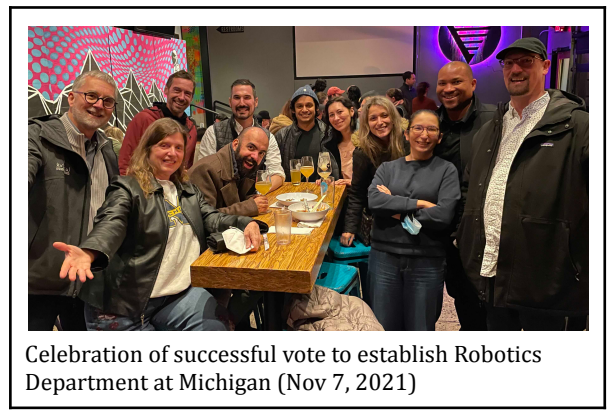

Celebration of successful vote to establish Robotics Department at Michigan (Nov 7, 2021)

The Faculty of the College of Engineering and the Michigan Association for State Universities has been tremendous in their support for Robotics and their willingness to take this step forward with an ambitious new degree program. We particularly appreciate Fred Terry, Dale Carr, Kevin Pipe, Betsy, Dodge, and Stacie Benison at the College Curriculum Committee for their patience and understanding through numerous CARF submissions and revisions, Nilton Renno at the College Executive Committee for crucial feedback to craft a compelling case for winning the approvals for the Robotics Department, and Kerri Wakefield, Mark Collyer, and Nick Gupta for helping us better understand what students need throughout their progress in Michigan Engineering.

Michigan Robotics has benefited greatly from wonderful staff support by our award-winning Student Services Team, including Denise Edmund, Virginia

Neal, Christina Hollis, and Kayla Dombrowski, who were critical to translating our ideas and collective dedication to students into reality.  Karen Revill in the ADUE Office deserves huge thanks for the endless coordination of schedules and meetings throughout this entire process.

Our biggest thanks must go to the students, staff, and faculty of Michigan Robotics.  It is you that has given life to our program.  Words cannot express how much we appreciate the dedication you give to make our community awesome.  There is so much that people in the Michigan Robotics community have given to build a truly special environment: your tireless efforts to develop and improve our courses, your long nights soldering and coding for course projects, your empathy and mentoring of younger students to help them along their journey, and your creativity and ingenuity that inspires people that walk through our building.  Michigan Robotics continually strives to earn your trust and confidence and live up to our values for Robotics with Respect.

We are also deeply grateful to the sponsors who have enabled our successful launch of the Robotics Undergraduate Program.  Funding from Ronald D. and Regina C. McNeil established the [McNeil Walking Lab](McNeil Walking Lab) that inspires students from the moment they walk into the Ford Motor Company Robotics Building.  Gifts from Roger Ehrenberg and Carin Levine Ehrenberg have provided enhanced staffing of undergraduate Instructional Aides for more meaningful student support in our courses and Distributed Teaching Collaboratives efforts.  Funding from the Toyota Research Institute, Amazon, and the Alfred P. Sloan Foundation has supported our early work to establish the Distributed Teaching Collaboratives model as a new approach to collaborative open-source course development with Historically Black Colleges and Universities.

The Michigan Robotics Undergraduate Program stands on the shoulders of giants.  As such, we must also pay homage to the groundbreaking robotics programs that paved the way for Michigan Robotics.  The foresight of Raj Reddy of the CMU Robotics Institute and Ruzena Bajcsy of the Penn GRASP

Lab remain the gold standard of foresight that we have always strived to meet. Henrik Christensen formerly at Georgia Tech and Maja Mataric at USC and played critical roles in catalyzing the academic ecosystem in the 2000s for new degree programs in robotics. Henrik's continued efforts with the CRA Computing Community Consortium catalyzed the growth of robotics into its discipline through the national [Robotics Roadmap](#) and the National Robotics Initiative. Our program takes considerable inspiration from the Robotics Engineering Department and undergraduate major created in 2007 and led by Michael Gennert at WPI – which was such an immense contribution to the advancement of academic robotics. The lead author must also thank Signe Redfield of the U.S. Naval Research Laboratory for her immense insights on what it means to create a new discipline and how an undergraduate curriculum defines an academic discipline.

As we launched this new robotics degree program, there are a number of special individuals that made especially significant contributions (and sacrifices) during our "startup mode." Creating and teaching fresh new introductory courses on a rapid timeline for real robots is not for the faint of heart. We had the great fortune to work with students, staff, and faculty who were able to rise to this special occasion: Maani Ghaffari, Dwane Joseph (Morehouse College), Tribhi Kathuria and the [Robotics 101](#) course staffs; Jana Pavlasek, Jasmine Jones (Berea College), Karthik Urs, Tom Gao, Brody Riopelle, Max Topping, and the [Robotics 102](#) course staffs; Ilya Kovalenko, Matt Romano, Derrick Yeo, Robin Fowler, and the [Engineering 100.850](#) course staffs; Lionel Robert, Joe Montgomery and the [Robotics 204](#) course staffs; Katie Skinner, Yves Nazon, Onur Bagoren, Chien Erh (Cynthia) Lin, Senthur Raj, Liz Olson, Anthony Opipari, and the many contributors to the launch of our 300-level Robotics courses; Madhav Achar, Grant Gibson, Eva Mungai, Wami Ogunbi for many contributions across our curriculum and impromptu robot demos; and our faculty and doctoral students who were willing to take on teaching overloads at critical points during our launch year – Anouck Girard for Self-driving, Leia Stirling for Probability and Statistics, and Anthony Opipari for Deep Learning for Robot Perception (check out [deeprob.org](#)).

| 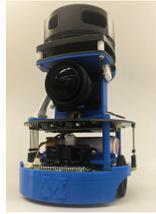 | 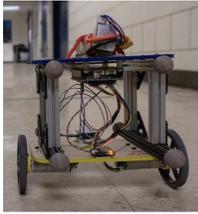 | 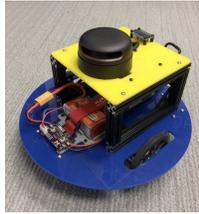 | 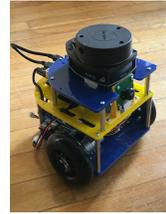 | 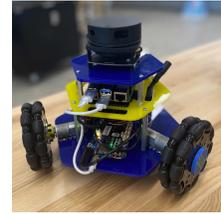 | 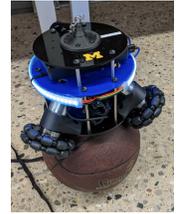 |
|---|---|---|---|---|---|
| MaeBot (2014) | BalanceBot (2016) | MBot (2017) | MBot Mini (2020) | MBot Omni (2021) | MBot BallBot (2022) |

The Michigan MBot family of mobile robots used throughout our undergraduate and graduate robotics curriculum.

Peter Gaskell, Abhishek Narula, Stanley Lewis, Cameron Kisailus, and Isaac Madhavaram were absolute heroes to get our fleet of over 200 MBot mobile robots ready for our first semester offering this Robotics Major.

These amazing contributors across Michigan Robotics are breaking the mold of what it means to be true leaders in robotics – as innovators who excel in both research and teaching – as scholars who balance theory and practice – and as thought leaders who understand the value of both people and ideas.

And, to those pioneering students who have taken these new robotics classes and declared Robotics as their major in our 2022-23 launch year …

Thank you for taking a chance on us! We will continue to blaze new trails and make history together!

# Preface from the Lead Author – Answering a Simple Question

In 2019, the Michigan Robotics Major started from a simple question –

What is the best major for a student to become a roboticist?

This question mystified me for years before joining Michigan Robotics in 2015. And, if you ask around in our field, you will likely find more opinions than seats in [Michigan Stadium](). But, the actual answer has been quite clear.

The best major to be a roboticist has been a quadruple major in Mechanical Engineering, Electrical Engineering, Computer Science, and Mathematics … with a minor in Psychology!

It would take a lifetime to learn everything that goes into our modern robots, as well as the possibilities for robot technology to shape our future as a society. Our goal should not be to try to pack all of this knowledge into four years – into a one-size-fits-all undergraduate major where survival of the fittest becomes the norm.

An robotics curriculum needs to provide the inspiration and foundation for how to think like a roboticist. In general, an undergraduate major defines the intellectual organization for its academic discipline to produce "people and ideas." In this spirit, we addressed the curricular challenge of how to both: 1) educate people to put ideas of the robotics discipline into practice and 2) endow them with the intellectual lens for creating new ideas that extend the frontiers of the robotics discipline.

Michigan Robotics has also strived to create a learning environment that is not simply an academic endeavor, but understands the real effect our classrooms and our research have on the world. Robotics, as a nascent discipline of innovation, has an imperative to lead in shaping an equitable future as society and automation converge. Our nation and our global community has a critical need to cultivate a thriving workforce of thoughtful and talented people. People who understand – using ***embodied intelligence*** as an organizing definition – how to make robots as machines that must sense, reason, act, and work with people to improve quality of life and productivity equitably across society.

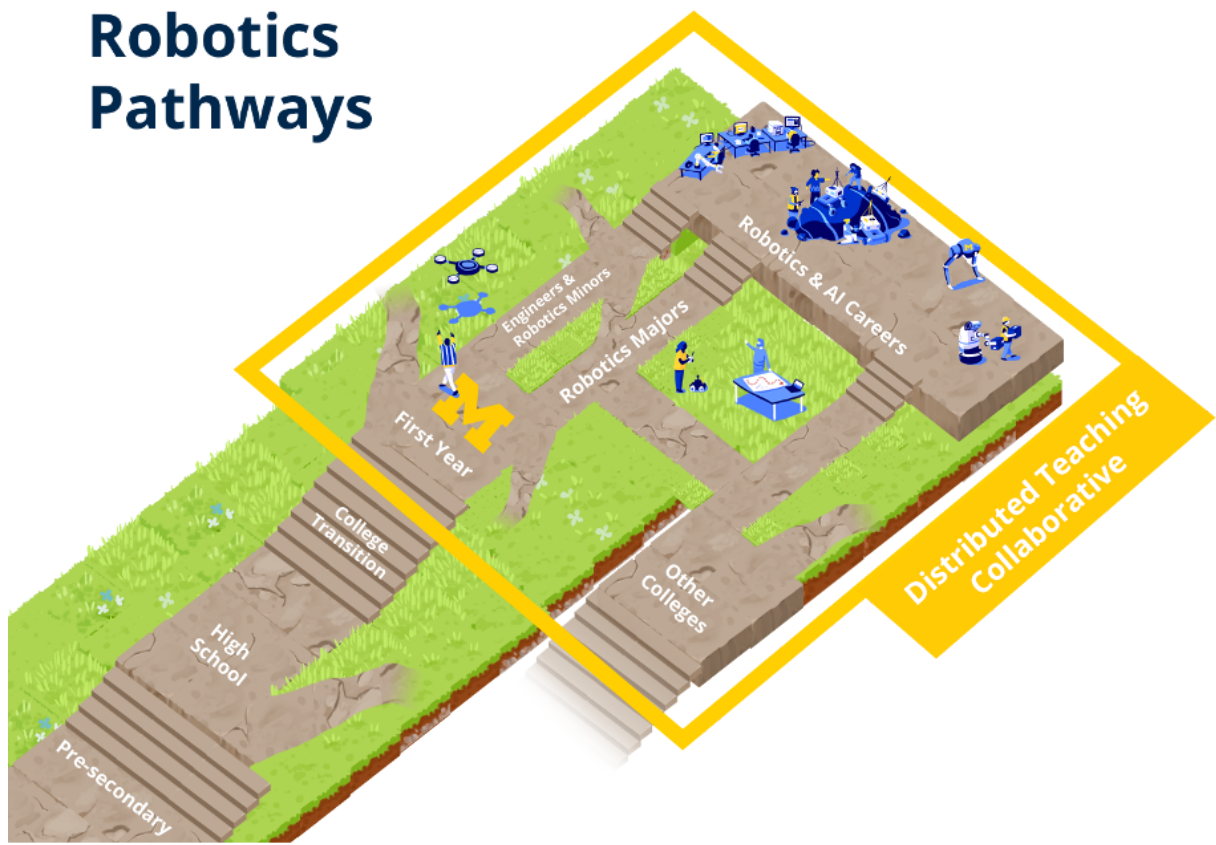

To meet this need, Michigan Robotics has created this Robotics Major and envisioned a larger Robotics Pathways model (illustrated above) to make robotics more meaningful and accessible across zip codes, time zones, and geographic boundaries – with the purpose of enabling successful and sustained careers in the robotics professions in a manner that reflects the diversity of our society. The Robotics Pathways builds on innovations introduced in our new Robotics Undergraduate Major, including:

- The Full Spectrum Robotics Introduction to "Inspire from Day 1" for students interested to pursue robotics from their first day on campus

  - The "Robotics Super Semester" that allows students to holistically experience embodied intelligence in one semester through three co-requisite courses for the "sense, reason, and act" of robotics

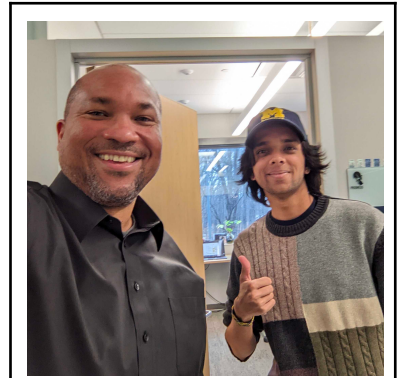

Adi Balaji – the first student to complete the "Robotics Super Semester" – with Prof. Jenkins (Fall 2022)

- Creating new sophomore and junior-level courses that elevate compelling topics, such as robot localization and mapping and robot kinematics and design, up from the graduate and senior levels.

- Bringing mathematics to life through the lens of robotics, such as leading with linear algebra for compelling applications and motivating the raison d'etre of calculus

- Improving pathways to graduate school from Minority Serving Institutions and Predominantly Undergraduate Institutions through collaborative open-source course development – which we call Distributed Teaching Collaboratives

- Enabling future generations to imagine themselves as roboticists through partnership with K-12 schools – such as our collaboration with the School at Marygrove in Detroit.

- Continual development of technical communications aptitudes throughout the curriculum, from introductory design through the capstone experience

- Embracing the spirit of experiential learning in Michigan Engineering through a flexible curriculum

- The Michigan MBot class of mobile robots as an affordable platform ($360 per robot) capable of full autonomous navigation for teaching robotics

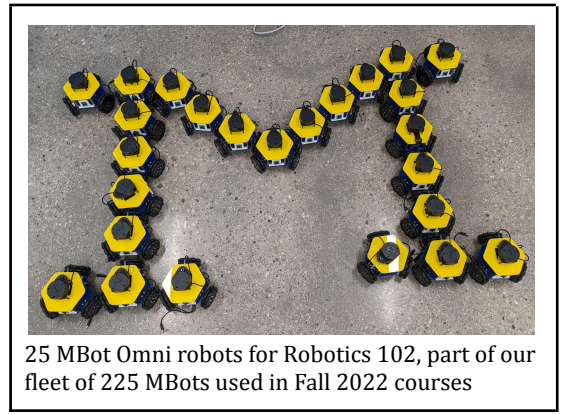

25 MBot Omni robots for Robotics 102, part of our fleet of 225 MBots used in Fall 2022 courses

- Working with our industrial partners to make the career fair feel more like "Star Trek" as an endless frontier of collaboration rather than "The Lord of the Flies" and its hypercompetitiveness.

More information about the [Michigan Robotics Undergraduate Program](#) is available from the [Michigan Robotics webpage](#). The inaugural [Robotics Undergraduate Program Guide](#) from 2022-23, as well as video from our [Fall 2022 Robotics Undergraduate Community Meeting](#), provide more specifics about the Robotics Undergraduate Program and curricular specifics in practice.

Through this Robotics Major, our aim is to realize equity and excellence in scholarship more than the values of Michigan Robotics, but also the calling of our discipline. We welcome continued discussion and collaboration on this opportunity to realize "Robotics with Respect" together as a community.

Thank you for taking time to listen to our ideas!

Odest Chadwicke Jenkins, Ph.D.
Professor of Robotics
Professor of Electrical Engineering and Computer Science
Inaugural Associate Chair for Undergraduate Education
Robotics Department
University of Michigan

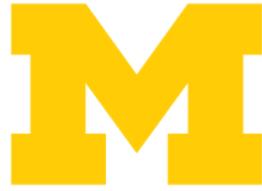

# Proposed new Bachelor of Science major in Robotics

February 18, 2022

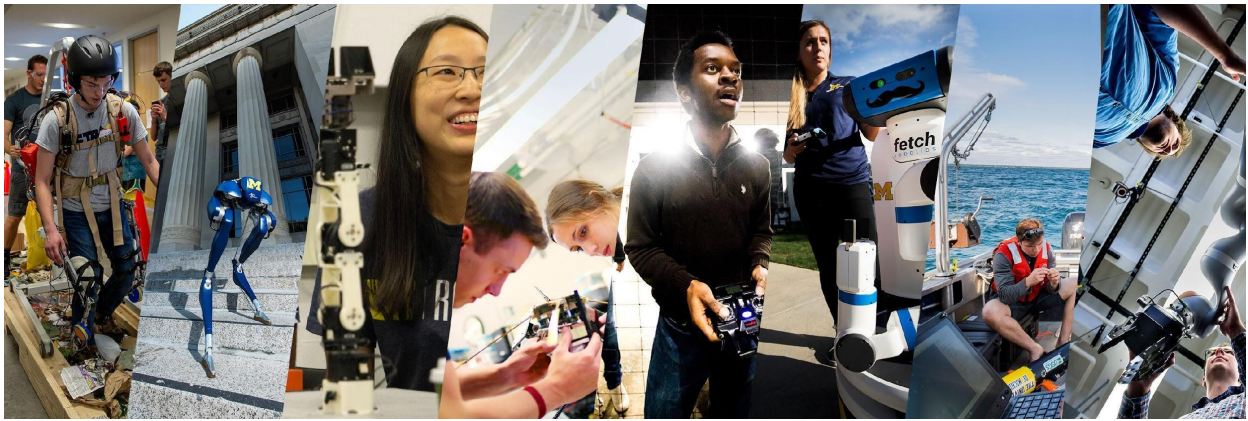

# Institutional Mission and Future Direction

Robotics, as defined by the National Science Foundation (NSF), is the "science of embodied intelligence," or put simply, the study of machines that sense, reason, and act in our physical world.

At the University of Michigan, robotics has risen to the level of its own discipline that is distinct from, though still intertwined with, traditional fields of study. In December, 2021, U-M regents made this official by approving the creation of the Department of Robotics–the first such department among top-ten engineering schools.

U-M Robotics has 30 core faculty from 12 departments. These faculty are leading experts from a variety of fields, such as aerospace, biomedical engineering, computer science, artificial intelligence or electrical engineering, and work on disparate applications, from drones to prosthetics to industrial manufacturing. But, they all must grapple with the fundamentals of robotics: sensing, reasoning, and acting.

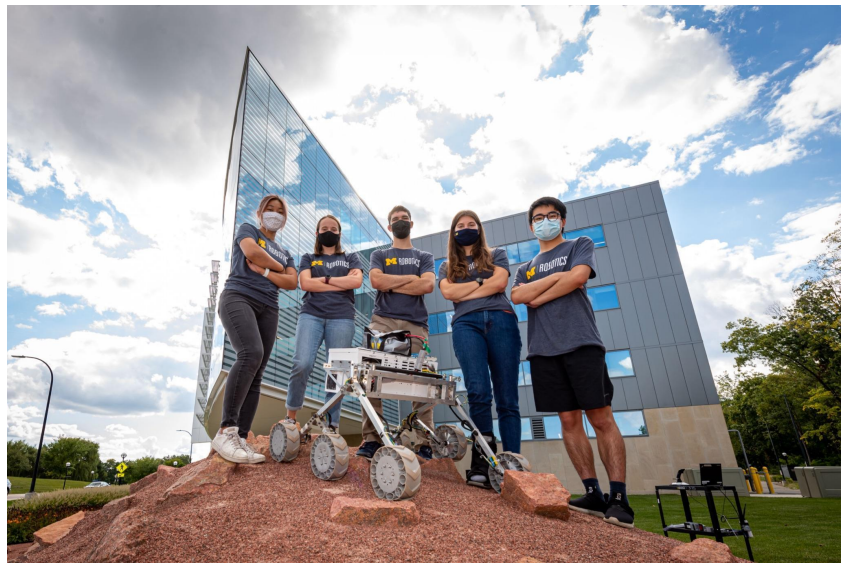

The new department will take shape from the existing Robotics Institute. The institute is headquartered in the recently completed $75 million, 134,000-square-foot Ford Motor Company Robotics Building—the world's most advanced dedicated robotics facility. The Robotics graduate program, which started in 2014, has grown into a Top-2 program with over 200 master's and doctoral students enrolled.

Many other schools have taken pride in advancing robotics education to its current state: Carnegie Mellon University and the University of Pennsylvania both began robotics institutes in 1979, which eventually led to the first graduate degree programs. The state of Michigan itself is uncommonly strong in robotics programs at the undergraduate level: Lake Superior State University was the first to offer an accredited undergraduate degree program in robotics engineering technology in 1985, and since then, Lawrence Tech, Michigan Tech, University of Detroit Mercy, and U-M Dearborn have started their own programs.

Now, in Ann Arbor, our proposed undergraduate program will be the first supported with the full resources of a dedicated department at a top public research university. Through its existing leadership in robotics, U-M is well poised to cultivate the leaders, innovators, and contributors who will tackle the needs of the 21st century.

Our aims with such a program are to:
- Meet growing demand from industry and the innovation ecosystem, which are urgently seeking robotics professionals
- Meet growing demand from students, who are increasingly excited to study robotics but must often wait until their third or fourth year of study for robotics courses
- Utilize our unique position to pilot innovations for educating and engaging students, both at U-M and through collaborative partnerships with other schools
- Define the discipline of robotics with excellence in equity and scholarship as top priorities on a level footing

Through the proposed program, roboticists will graduate in four years, instead of the six years that it requires to earn both an engineering Bachelors of Science and robotics Master's degree. Core robotics topics will be elevated from the graduate and senior levels into Years 2-3 of the undergraduate experience. These elevated topics include [Simultaneous Localization and Mapping](#) that is essential for modern autonomous vehicles and Forward and Inverse Kinematics needed to control robot arms and mobile manipulators, such as the Fetch (pictured right).

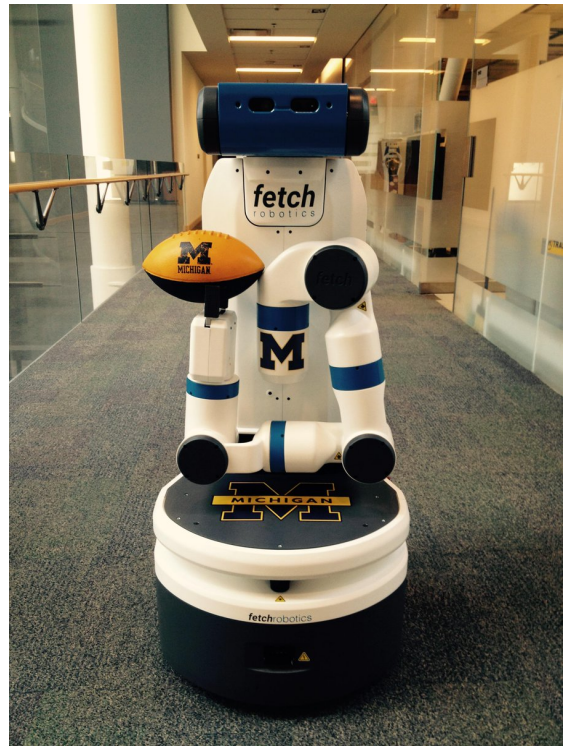

Building on lessons learned from existing robotics programs, the proposed robotics degree aims to provide a clear intellectual organization of the discipline of robotics through our undergraduate curriculum. This curriculum will both lead in how we develop new scholarship in robotics and educate students capable of the practice and advancement of robotics. Michigan also has the unique opportunity to define the discipline of robotics with a priority on both equity and excellence. Towards this end, the proposed curriculum will establish partnerships for distributed teaching collaboratives, especially with Historically Black Colleges and Universities (HBCUs) and other Minority Serving Institutions (MSIs).

Our proposed undergraduate robotics major will meet emerging needs, define the discipline, and innovate for equity.  The remainder of this document describes a unique in-person on-campus undergraduate experience designed to inspire through a modern approach to robotics and engage emerging scholars across our community, our nation, and our world.

# Need and Rationale

Based on findings from the Robotics Department Planning Committee of 2021, which commissioned a market research study, we are confident in the need for the proposed undergraduate program based on the below factors.

**The academy needs to keep pace with the growing demand for jobs, technology, and innovation**: The national and global technology landscape is shifting, and U-M is poised to meet this challenge. The US Bureau of Labor Statistics recently reported that the annual demand for qualified robotics professionals grew by over 13% in 2018 alone. In addition, up to 80% of US industrial employers are facing difficulties filling vacancies for highly skilled technical professionals, including robotics, computer vision, artificial intelligence, and motion control.[1]

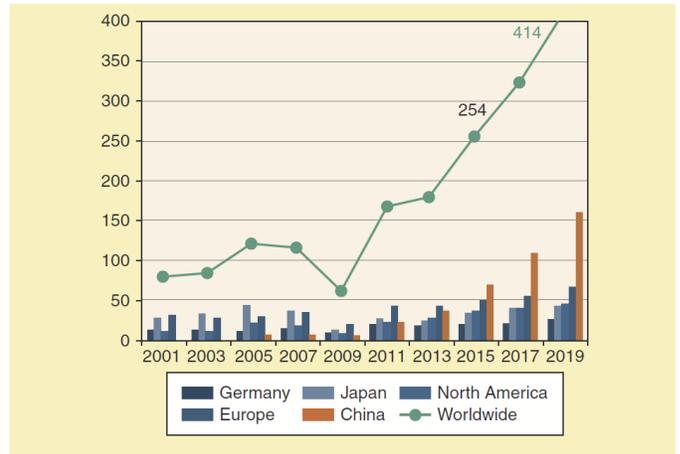

**Figure 1.** The industrial robot shipments worldwide in thousands of units (2016+ estimated).

These statistics are underscored by a growing market. The global industrial and service robotics markets are expected to grow by over 20% year over year (Compound Annual Growth Rate), resulting in a total market value of nearly $210 billion by 2025.[2] The writing is on the wall. The confluence of a gap in skilled workers and a growing market demonstrates the need to rethink technical higher education. That is, a new program is needed that i) provides interdisciplinary training in robotics curated to meet current and future technology demands of the robotics workplace, and ii) provides a college major, a clear designation of skills that will enable graduates to compete for the most promising and highly skilled jobs.

**Students are searching for universities that offer robotics undergraduate majors and minors**: There is an explosion of interest in robotics-related activities from young people of all ages, and a robotics department will draw these students. Middle and high school students interested in robotics have historically been unable to learn the desired technical content in school, and have instead turned to afterschool programs. FIRST Robotics—a league where teams compete with custom-built robots completing predefined tasks—was developed to meet this growing need. Over the past decade, the number of FIRST Robotics teams has grown to serve over 679,000 student participants in over 100 countries, including 508 in Michigan. A similar program, VEX Robotics, has expanded to over 50 countries over the same period. These programs have a direct and meaningful impact on participating students. A recent study found that over 95% of students participating in VEX Robotics desired to learn more about robotics and engineering in their future education.[3] While the programs are impressive, they represent a groundswell of interest from younger generations. These students have already demonstrated the lengths they will go to study robotics by pursuing these experiences in their free time outside conventional education. Without a clear and strong emphasis on robotics—namely, a Robotics department with an undergraduate major—U-M may struggle to attract these promising students.

---

[1] Shmatko, N., & Volkova, G. (2020). Bridging the skill gap in robotics: Global and national environment.
[2] Nexus report, pg. 4

**Demand at U-M for Robotics Undergraduate Education**: Market research shows that as many as 40 percent of current students expressed interest in a U-M Robotics major or minor, while 25 percent of students who were accepted to U-M but ultimately enrolled elsewhere indicated that they might have decided differently if U-M currently offered a Robotics degree program. On one level, this underscores the need, even the urgency, for launching a full undergraduate Robotics curriculum. It also points to some important decisions the College and department will have to make if demand is as high or higher than projected, including the possible need to cap enrollment in a way that is equitable, meeting Robotics' and Engineering's commitment to diversity.

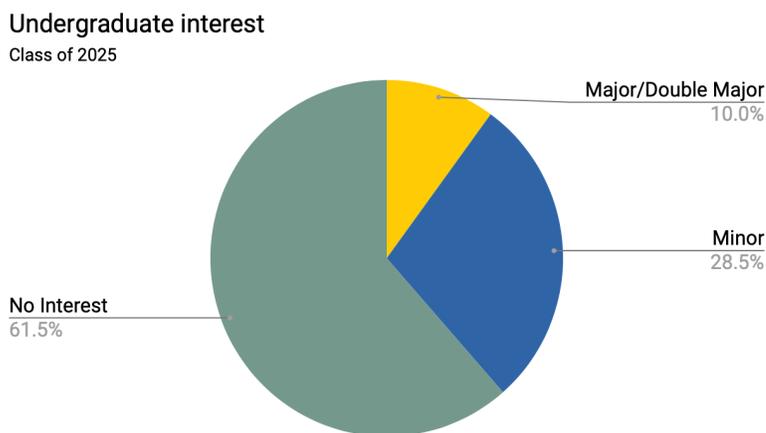

# Background & Values

The U-M Robotics Undergraduate Program has been formed for innovation, building on the foundation of U-M Engineering and furthering the ethos of "[Robotics with Respect](#)." U-M Robotics aims to meet undergraduate intellectual needs emerging in the 21st Century by inspiring students from their first day on campus and cultivating equal opportunities for a diverse world. Indicators of student interest, both formal and anecdotal, show high interest among undergraduates for a Robotics major. Our society has a growing and unmet demand for people skilled in robotics, as well as artificial intelligence. Michigan is well poised to lead the evolution of higher education to meet these challenges, affect positive systemic change, and prepare future generations for a highly dynamic innovation ecosystem.

Robotics faculty have taken the opportunity of a new department to rethink how to meet the modern needs for producing excellence in both equal opportunity and leading scholarship. The values of U-M Robotics foster a culture of empathy, wellness, and growth to help our students develop into good citizens of our profession. Our efforts to design a curriculum that favors a student's success being determined by their intellectual ability and drive instead of where one attended high school are reflected throughout the entire undergraduate curriculum. U-M Robotics aims to cultivate leaders in academic excellence and support personal development at all career stages, promote innovation in undergraduate and graduate programs, and engagement with partners in K-12 education, MSIs, industry, and the public sector.

---

[3] Hendricks, C.C., Alemdar, M. and Ogletree, T., 2012. The impact of participation in vex robotics competition on middle and high school students' interest in pursuing stem studies and stem-related careers. American Society for Engineering Education.

The first trial of this work was our pilot course offering of ROB 101: *Computational Linear Algebra*, which won a [2021 Provost's Teaching Innovation Prize](). This course reimagines the way mathematics is introduced to first-semester engineering undergraduates by integrating math with programming to allow engineering projects at the "scale of life". Examples from this course include building [3D maps for robot navigation from LiDAR data]() and controlling a planar [model of a Segway using optimization](). This curricular innovation will allow students to have a palpable understanding that computation and mathematics are their friends instead of hoops to be jumped through on the way to a degree. They will be prepared to experience computation, mathematics, and science as tools that allow them to construct better, safer, more reliable machines. We will have broken down the stove-piping of knowledge by integrating knowledge acquisition with its use.

We also partnered with Morehouse College to concurrently teach students the course material, participating in remote lectures, discussions, and office hours. We are now extending this to create a Distributed Teaching Collaborative with MSIs, with explicit planning in course offerings to include remote participation of students from HBCUs, collaborating with them on a summer camp for high school students run through Morehouse, and placing course content online—not just the videos of the lectures, but actual course notes and supporting material.

In Fall 2021, Robotics 102: *Introduction to AI and Programming* expanded our Distributed Teaching Collaboratives effort to partner with Berea College. Berea College primarily serves students in the Appalachian region, providing upward mobility to its students as a tuition-free full-participation work-study institution.

U-M Robotics faculty actively participate in demographic-specific conferences and venues (e.g., NSBE, SWE, SHPE) to engage students who may not have otherwise considered applying to U-M. All current faculty and staff who hold appointments in the Robotics Institute have completed the Change it Up! To Stop Anti-Black Racism training module. Robotics faculty have developed an 11th grade course in cooperation with Detroit metro schools and, as cited above, founded Distributed Teaching Collaboratives with HBCUs, which are specifically designed to realize the potential of underserved communities to contribute to–and participate in–the field of robotics. U-M Robotics has also invited Ford Motor Company to assist with these efforts, introducing Ford employees who have the expertise of putting robust engineering on roads to help contribute to robotics education.

All these efforts align with the U-M College of Engineering's (CoE) mission of equipping graduates with the skills needed to design equity-centered solutions to society's challenges. The efforts also align with the greater university's direction of serving the people of Michigan, and beyond, by creating leaders who can apply their knowledge to enrich the future.

# Defining the Discipline of Robotics: Sense-Reason-Act

In building the discipline of robotics, our principles organize around the National Science Foundation's [definition of embodied intelligence](#) and the idea that robots must sense, reason, and act. Robots must take in information to perceive the world around them, plan out how to achieve a goal given their perception of the world, and then carry out that plan through controlling its motors and interacting with their physical environment. The study of robotics requires knowledge of how a robotic system sees its surroundings, how and what it can move around, and how doing so will affect its surroundings. While specific research questions in robotics can become much more technical, at the top-level, they fall within the broad categories of sensing, reasoning, or acting. In alignment with the NSF's definition, the learning objectives for our proposed undergraduate program are organized around these principles of embodied intelligence.

Our educational objectives are formed to equip students with the foundational skills and knowledge they need to thrive as competent roboticists and enable them to continuously grow throughout their careers. We further require a robotics degree program to allow students the flexibility to specialize in different areas within robotics. In addition, our learning objectives will remain consistent across the major, and in complement and alignment with the educational objectives of the College of Engineering.

Given these requirements, we propose that the discipline of robotics at the undergraduate level is organized by the following learning objectives and concepts:

- Foundational Learning Objectives:
    - Computational fluency to express ideas through coding – *"coding is believing"*
    - Design, Maker, and Shop competency for realizing systems that can move in the physical world
    - Linear algebra to structure, process, and manipulate data at scale
    - Human and social dynamics for working in teams to develop solutions for people
    - Technical communication and project management to understand and meet the needs of stakeholders in a professional and conscientious manner
- Intermediate Robotics Concepts for sound understanding in at least 3 of the following core areas of robotics:
    - Localization and Mapping for autonomous navigation
    - Kinematics, Planning, and Simulation for control of robot manipulators
    - Mechanical Design and Dynamics for fabrication of physical robot systems
    - Electrical Sensors and Signals for design of robot circuitry and embedded systems
    - Human-Robot Interaction for measuring and quantifying robot efficacy
- Advances and applications in Robotics: exposure to the diversity of topics for students to reach and extend the cutting-edge of knowledge, which include
    - [Bipedal Locomotion](#)
    - [Neurorobotics](#)
    - [Marine Robotics](#)
    - [Experimental Unmanned Aerial Vehicles](#)

What makes the discipline of robotics different from other disciplines is the specific combination and depth of knowledge from these intersecting topics. There is, of course, an overlap in the concepts and skills needed with other programs, such as mechanical engineering and electrical engineering. However, core requirements of those overlapping programs, like comprehensive study of thermodynamics or discrete math, remain elective-level objectives in robotics. Similarly, core requirements in robotics, topics that enable an embodied intelligence to sense, reason, and act, are elective in other fields.

By meeting the learning objectives above, a robotics major will thrive in a market demanding well-rounded, experienced, and knowledgeable roboticists. The learning objectives center upon the principles of sensing, reasoning, and acting, and include topics in electronics, mechanisms, computation, mathematical foundations, and human-robot interaction. Instantiated around the philosophy of providing students with a full spectrum robotics experience, the proposed curriculum (described in the next section) gives students the ability to focus on topics of specific interest in depth, complemented by exposure to the disciplinary breadth of engineering at the intersection of robotics and the many intersecting disciplines across the academy.

# Full Spectrum Robotics Curriculum

To meet the learning objectives above, we propose the Full Spectrum Robotics Curriculum and discuss its major features as an in-person on-campus undergraduate experience. An undergraduate curriculum provides the embodiment of a discipline, for both organizing and disseminating its intellectual foundations. In this spirit, the Full Spectrum Robotics Curriculum gives shape to the most important ideas in robotics and acculturates students to practices of the field, situating them for continued professional success. This curriculum accommodates the learning needs of students who strive to be roboticists and improve the human condition.

## Full Spectrum Introductory Sequence: Sense, Reason, Act

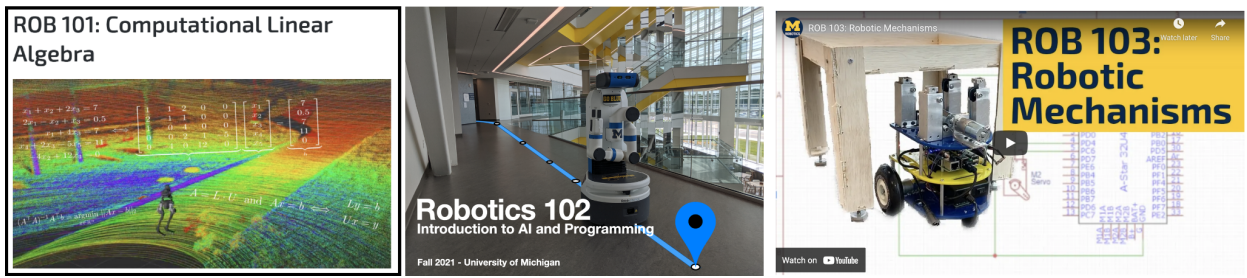

One unique and critical component of the proposed Full Spectrum Robotics Curriculum is the introductory class sequence. This introduction is designed to align with the first-year experience of the College of Engineering along with innovations to inspire students that are eager to engage with robotics. Our new robotics 100-level courses offer a flexible and immersive introductory experience into the discipline of robotics. This introductory pathway will acquaint students with the foundations for modern engineering through the lens of autonomous navigation for mobile robots. In compliance with the educational philosophy of U-M Engineering, our 100-level course offerings also offer a seamless pathway into the robotics major for students who have not taken the robotics-specific 100-level classes. Thus, students who decide to study robotics at U-M early in their academic career will have that opportunity, while students who make that decision later will not be penalized.

Programmatically, incoming students are admitted to the College (not a specific department), and typically declare a major in their second or third term. During their initial terms, students fulfill the CoE core requirements by taking general courses that provide the foundations needed for modern engineers. As shown in the table below, these foundations begin with Engineering 100 and Engineering 101, along with an emphasis on mathematical preparation in Linear Algebra.

Our curriculum provides robotics-focused versions of these foundational engineering courses, as ROB 102: *Introduction to AI and Programming*, and Engineering 100.850/ROB 103: *Robotic Mechanisms*. Similar in aims to Engineering 101, ROB 102 provides an introductory experience to computational thinking and programming in preparation for data structures courses. ROB 102 conceptualizes computing as graph and graph algorithms that are grounded in various approaches to path planning for autonomous navigation by mobile robots. In alignment with sections of Engineering 100, ROB 103 is an introductory real-world design experience to build and control a mobile robot inspired by warehouse robots used in supply chain logistics. ROB 103 familiarizes students with shop facilities for making physical systems and introduces students to low-level controls and embedded programming.

| Foundational Engineering Courses | Current course | Complementary ROB course | Content tied to robotics concepts |
|---|---|---|---|
| Introduction to Engineering | ENGR 100 | ENGR 100.850 (ROB 103) | design experience focused specifically on robots |
| Computation and Programming | ENGR 101 | ROB 102 | introductory computer programming through AI and graph search for autonomous navigation |
| Linear Algebra | Math 214 and 217 | ROB 101 | representing spatial systems for 3D mapping; large linear systems of equations for bipedal control |

Our introductory curriculum also introduces innovations to the mathematical pathways to both prepare and motivate students for the analytical needs of modern engineering. Instead of seeing the "rules of calculus" as their introduction to college mathematics, students will experience the "raison d'etre of mathematics in engineering", in other words, why we employ it so heavily in engineering in the first place.

ROB 101: *Computational Linear Algebra* is suited to complement traditional and more theoretical linear algebra courses (such as Math 214 and 217) while giving students sufficient preparation for intermediate and upper level courses in Engineering. ROB 101 provides the mathematical concepts for representing spatial systems, such as for 3D mapping, and large linear systems of equations, such as for bipedal control.

## Human-Robot Gateway: Elevating the Human Dimension

ROB 204: *Human-Robot Systems* serves as the gateway course into the intermediate level for robotics majors. It introduces students to the human dimensions of robotics, including introducing human-information processing models to support robotic system design and real-world integration. Robotics 204 presents both the usability of robotics and the structure, the ethical considerations for robotic systems, and workflow for human-centered design to inform project management. Technical communication is a featured element of Robotics 204 and serves to provide continued touchpoints into the Tech Comm curriculum throughout the undergraduate experience.

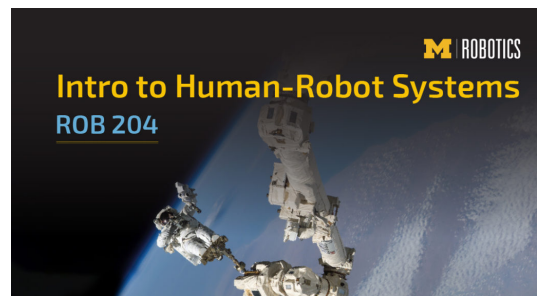

Although this robotics-focused sequence provides students the opportunity to delve more deeply into robotics early in their degree, it does not compromise their ability to declare their major. Students who take different introductory pathways (e.g., ENGR 100, ENGR 101, and Math 214) will be accepted into ROB 204 (as the "gateway" course to the robotics major), so that not taking the ROB 100-level sequence does not preclude students from majoring in robotics. Conversely, taking the ROB 100-level sequence of classes would not hinder students who decide to declare a non-robotics major within the College of Engineering.

## Intermediate Level Customization: Breadth with Depth

A unique feature of the Full Spectrum Robotics Curriculum is the elevation of core robotics topics into Years 2-3 of the undergraduate experience. Currently, students must wait until their senior year or a graduate program to take courses essential to robotics. The 300-level courses in the proposed degree program fulfill this unmet need by providing a firm grounding in the core dimensions of robotics while accommodating the diverse range of pathways across the discipline of robotics.

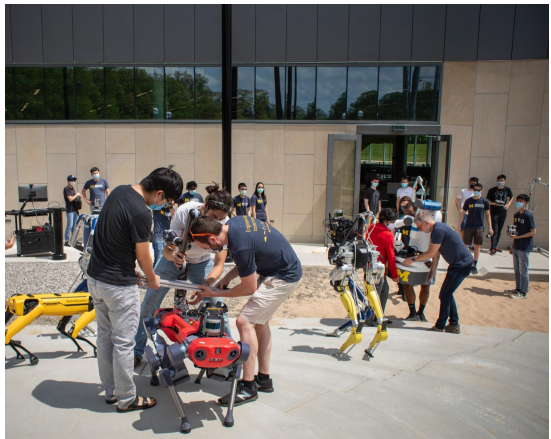

Students majoring in Robotics are required to take at least three of the 300-level Robotics courses, with the flexibility to take all five. This structure of the intermediate level of the degree allows students to customize for either breadth or depth in robotics, taking inspiration for the [threaded approach to curriculum design for computing](#).[4] Students can begin to specialize at the intermediate level while gaining sufficient overlap with the core of the discipline. More specifically, Robotics majors must have touch points into at least two of the three major groupings of core topics:

- Hardware-focused mechatronics
  - ROB 310: *Robot Signals and Sensors*
  - ROB 311: *Build Robots and Make Them Move*
- Computing-focused autonomy
  - ROB 320: *Robot Operating Systems*
  - ROB 330: *Localization, Mapping, and Navigation*
- Empiricism-focused human-robot interaction
  - ROB 340: *Human-Robot Interaction*

A student striving to be a well-rounded roboticist can take all five of these courses, where two can serve as upper-level electives. A student looking to specialize can complement their learning at the intermediate level with upper-level elective courses. Both breadth and depth pathways through the intermediate level provide a suitable foundation for further exploration in upper level courses, as well as continued learning into other core areas of robotics.

---

[4] Furst, Isbell, Guzdial,"[Threads™: How to restructure a computer science curriculum for a flat world](#)"

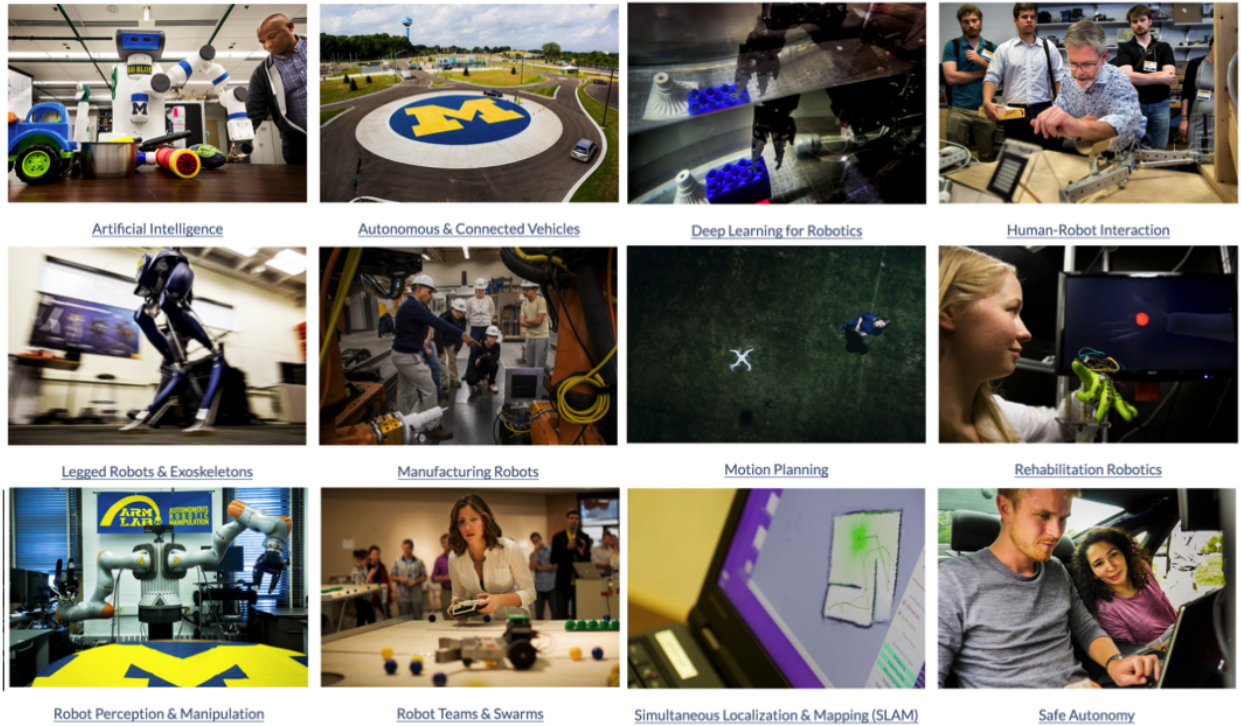

## Upper-Level Electives: Pushing the Envelope of the Possible

Once students reach the 400-level, they will be able to choose from a wide array of upper-level electives, allowing them to customize their degree plan to align with their personal interests and skills. These robotics upper-level courses will draw upon the areas of expertise (pictured above) of the U-M Robotics faculty. The flexibility of the upper-level electives gives students the opportunity to explore diverse topics and develop interests and skills that will prepare them for active contributions within the field of robotics. More interestingly, this Robotics degree will enable the University of Michigan to lead the way in offering courses unlike anything that has been available to undergraduates previously. Bipedal Locomotion, Neurorobotics, Marine Robotics, Experimental Unmanned Aerial Vehicles are just a few of the possibilities ready to be tapped into given the world-class robotics expertise on campus.

## Capstone Experience: Professionalism and Communication

Capstone experiences in Year 4 of the Full Spectrum Robotics Curriculum is designed to help students grow into engineers with the professional skills to understand and meet the needs of people. Learning from structures at peer institutions, the Robotics capstone is a two-course sequence of two new courses: TCHNCLCM 350: *Technical Communication for Robotics* and ROB 450: *Robotics Capstone*.  TCHNCLCM 350 will guide students through the process of client-driven project management and the necessary technical communication skills for such processes to be effective. Projects in this course are intended to be relatively straightforward from a technical perspective to enable effective process, communication, and teamwork strategies to be learned. These strategies are then put into practice in earnest the following semester in ROB 450 with participating external clients. This capstone structure was forged through collaboration with the Program in Technical Communications.

## Advising: Localizing the Sense of Connection

This curricular design will be complemented by an advising structure that aims to give students insights into possible professional pathways beyond graduation. Further, thoughtful advising that connects with students and their interests as individuals can combat one of the largest factors preventing diversity: *isolation*. Towards this end, our approach to advising will start from the premise of presenting students with possible professional pathways beyond their completion of a major in Robotics. Students will be able to envision what pathway is best for them, and work with their advisor to find a degree-satisfying course plan through the Robotics curriculum that will help them achieve their aspirations. This purpose-driven approach to advising will hopefully allow us to cultivate smaller and more localized communities of support in what will likely grow into a sizable degree program. Our efforts for advising are complemented by the advisory activities that occur in Engineering 110 and programs such as M-Stem, CSP, and Wolverine Pathways.

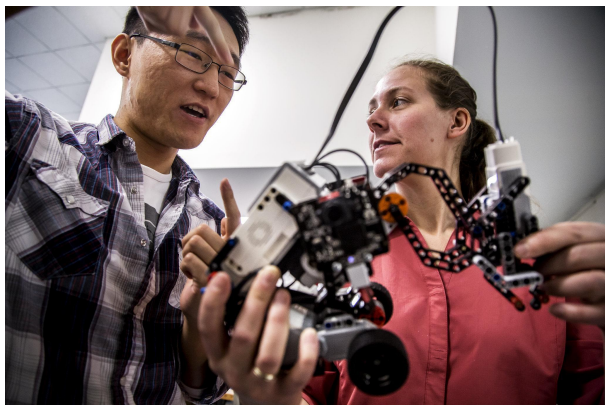 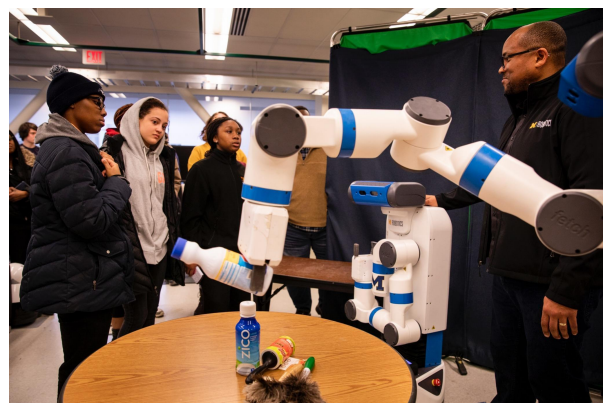

# Internal Status of Proposal

The new major has been approved by the College of Engineering through a majority vote of the faculty.

We aim to accept the first declared majors in Fall 2022.

# Related Programs

Although robotics is a thriving discipline, there are few undergraduate robotics programs outside of Michigan. Most universities with robotics programs offer graduate level studies with limited or no undergraduate components. When undergraduate programs are present, they are frequently limited to a minor, concentration, or certificate.

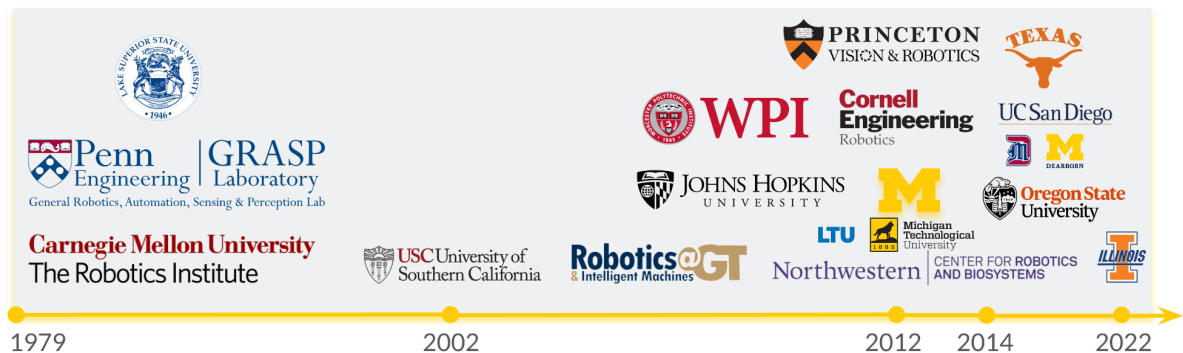

*A growing variety of institutes, centers, and labs mainly offer MS and PhD degrees. Undergraduate programs, besides several Michigan schools and WPI, are limited to dual majors, minors, or certificates.*

However, Worcester Polytechnic Institute offers an example of an established undergraduate robotics engineering program. The program, launched in 2007, has a focus on hands-on project work. Students design and control robot behavior while learning the core disciplines of Computer Science, Electrical & Computer Engineering, and Mechanical Engineering. The program complements the technical knowledge with courses in social implications and entrepreneurship to broaden student understanding of how robotics integrates with both society and industry.

| UNDERGRADUATE DEGREES IN ROBOTICS OUTSIDE MICHIGAN ||||
| --- | --- | --- | --- |
| UNIVERSITY | DEPARTMENT | DEGREE NAME | PROGRAM INFORMATION |
| Carnegie Mellon | Robotics Institute | Additional Major in Robotics | ● Designed for students who want to explore the field more than is possible through the Minor in Robotics<br>● Top 10 US News & World Report |

| | | | Engineering program |
|---|---|---|---|
| [Purdue University](#) | Manufacturing Engineering Technology | Robotics Engineering Technology, BS | ● Accept 12 high school credits from Project Lead The Way<br>● 1 of 3 majors offered in the Purdue Polytechnic Institute |
| [Rochester Institute of Technology](#) | Department of Manufacturing and Mechanical Engineering Technology | Robotics and Manufacturing Engineering Technology, BS | ● Four Co-op blocks required<br>● Curriculum: The first year focuses on writing courses<br>● Industry focus: Automotive, Defense, Manufacturing |
| [UC Santa Cruz](#) | Electrical & Computer Engineering | Robotics Engineering, BS | ● Capstone project required<br>● Leverages Computer Engineering's research and courses |
| [United States Naval Academy](#) | Weapons, Robotics and Control Engineering | Robotics and Control Engineering, BS | ● 1-year team based capstone project |
| [Worcester Polytechnic Institute](#) | Robotics Engineering | Robotics Engineering, BS | ● First BS in Robotics Engineering in the nation<br>● Project based learning<br>● Curriculum differentiators: entrepreneurship & social implications |

| ROBOTICS UNDERGRADUATE DEGREES IN THE STATE OF MICHIGAN | | | |
|---|---|---|---|
| **UNIVERSITY** | **ACADEMIC UNIT** | **DEGREE NAME** | **PROGRAM INFORMATION** |
| [Lake Superior State University](#) | School of Engineering & Technology | Robotics Engineering, BS | ● Also offer robotics minor |
| [Lawrence Tech University](#) | Mechanical, Robotics, and Industrial Engineering | Robotics Engineering, BS | ● Mentoring program for middle and high school students<br>● Internship & co-op opportunities |
| [Michigan Tech](#) | Electrical & Computer Engineering | Robotics Engineering, BS | ● Engineering design path: Robotic Systems, Aerospace, Blue Marble Security, or Wireless Communication<br>● Option to complete a year-long, industry-sponsored project centered on robotics with fellow ECE students through the Senior Design program |

| | | Robotics and Mechatronics Systems Engineering with a Concentration in Electrical Engineering | ● 3 paid co-op experiences<br>● 100% employment after graduation<br>● Mechatronics Systems is a career focus area |
|---|---|---|---|
| [University of Detroit Mercy](#) | Engineering and Computer Science | | |
| [University of Michigan - Dearborn](#) | Electrical and Computer Engineering | Robotics Engineering, BSE | ● Areas of focus electrical, computer, mechanical, automotive, and robotics |

Outside of bachelor's degrees, there are several community colleges with two-year robotics programs culminating in associate degrees. Many offer co-op programs with local businesses as well as partnerships with universities allowing associate degree graduates to pursue an undergraduate degree in robotics.

In comparison to these similar programs, the proposed U-M program shares the hands-on approach that many of these programs have. Offering students the opportunity to build hardware and code real projects leverages the student excitement toward the field while creating an avenue to teach the core engineering concepts required for a robust understanding.

Building on this, U-M's College of Engineering offers a broad base of courses with which to pair the core robotics knowledge, and U-M's other schools and colleges can offer complementary and nationally-recognized education in medicine, social sciences, biological sciences, public health, art, architecture, business, or any number of threads a student may want to pursue.

Many undergraduate programs currently offered are based out of electrical or mechanical engineering programs, and grant robotics engineering or mechatronics degrees. With a separate Robotics department and the full offerings of the University available, we can create a more broadly-defined robotics major that incorporates topics such as human-robot interaction and artificial intelligence that will further prepare students for advanced careers.

There is also a thriving student project team community in U-M's College of Engineering. Many of these team members participated in FIRST Robotics prior to coming to U-M, and desire continuing in an activity with similar camaraderie, engineering challenges, and competition. Teams at U-M include those that work on autonomous submarines, boats, aerial vehicles, and rovers, as well as teams that work on prostheses and exoskeletons. These teams are supported with shops and spaces where they can design, create, and test.

The proposed program will also differentiate itself in the delivery of math. The program's first pilot course, ROB 101 Computational Linear Algebra, elevates what is typically a higher level math course to first-year students, even before they might take calculus. Linear algebra is crucial in robotics, and allows students to begin deeper study of sensing, reasoning, and acting.

As discussed in the Rationale and Projected Enrollment sections, increasingly high industry demand and student interest necessitate the expansion of robotics programs to prepare larger numbers of highly skilled roboticists. The undergraduate robotics major will address a critical gap in educational options for undergraduate students interested in pursuing robotics.

| PROGRAM COMPARISON | | | | | |
|---|---|---|---|---|---|
| INSTITUTION | GRADUATE PROGRAM | UNDERGRAD MAJOR | UG AS DUAL MAJOR ONLY | UG MINOR | UG AS CERTIFICATE |
| Carnegie Mellon University | X | | X | X | |
| University of Pennsylvania | X | | | | |
| University of Southern California | X | | | | |
| Georgia Tech | X | | | | |
| Johns Hopkins University | X | | | X | |
| Oregon State University | X | | | | |
| Worcester Polytechnic Institute | X | X | | X | Certificate also available |
| Northwestern University | X | Concentration only | | | |
| Cornell | X | | | X | |
| University of Texas at Austin | Certificate only | | | | |
| University of Illinois | MS starting in 2022 | | | | |
| Lake Superior State University | | X | | X | |
| U-M Dearborn | | X | | | |
| Lawrence Technological University | | X | | | |
| Michigan Tech | | X | | | |
| University of Detroit Mercy | | X | | | |

# Curriculum and Degree Requirements

Below is a proposed set of course requirements for a Major in Robotics at the University of Michigan in accordance with the [Core Requirements for Undergraduate Programs in the](#) [College of Engineering](#).

## U-M Robotics Undergraduate Program Requirements for Majors

### Major Declaration Requirements
To declare a major in Robotics, a student must be a College of Engineering student and:
1. Have completed at least one full term at UM Ann Arbor
2. Have an overall UM GPA of 2.0 or better in courses taken at the UM Ann Arbor campus and be in good standing
3. Have completed or earned credit by exam or transfer for at least one course in each of these categories:
    a. Introductory Linear Algebra (e.g. Robotics 101 or Math 214)
    b. Introductory Calculus (e.g. Math 115, 116 or 156)
    c. Calculus-based physics lectures (e.g. Physics 140 or 160)
    d. Required introductory Engineering (Robotics 103 or Engineering 100, and Robotics 102 or Engineering 101)
    e. Teamwork in Robotics: ROB 204* Human-Robot Systems

### College of Engineering Core Program Requirements
1. Full Spectrum Robotics Introduction
    a. Introduction to Engineering: ROB 103 Robotics Mechanisms or Engineering 101
    b. Computational Thinking: ROB 102 Introduction to AI and Programming or Engineering 101 or introductory programming equivalent)
    c. Linear Algebra: ROB 101 Computational Linear Algebra or Math 214 or Math 217 or Math 417 or Math 419
2. Calculus requirements
    a. Introductory Calculus: Math 115 or Math 120 (AP); and Math 116 or Math 121 (AP)
    b. Intermediate Calculus: Math 215 or Math 216
3. Physics 140/141 and Physics 240/241
4. Chemistry 130 and 125/126*
5. Intellectual Breadth (16 credits as specified by the [College of Engineering Core Requirements Bulletin](#))
6. General Electives (15 credits)
    a. 15 credits are "required"; College degrees require 128 total credits, and more or fewer GE credits may be needed to achieve this total depending on individual factors in a student's record.

### Robotics in Engineering Program Requirements
1. Teamwork in Robotics: ROB 204 Human-Robot Systems

2. Robotics Core: at least three of the following courses:
    a. ROB 310 Robot Sensors and Signals
    b. ROB 311 Build Robots and Make Them Move
    c. ROB 320 Robot Operating Systems
    d. ROB 330 Localization, Mapping, and Navigation
    e. ROB 340 Human-Robot Interaction
3. Discipline Breadth: at least one approved course from all of the following areas:
    a. Data Structures and Programming: EECS 280
    b. Probability, Statistics, and Visualization: IOE 265 or EECS 301
    c. Electronics and Circuits: EECS 215 or EECS 270 or BME 211
    d. Kinematics and Dynamics: ME 240 or ME 360

4. Discipline Depth: one course from the following list (or approved by the Robotics Undergraduate Committee), such as IOE 333, AERO 201, EECS 373, NAME 270, EECS 281, EECS 370, EECS 216, EECS 351, MSE 220
5. Technical Electives: a minimum of 20 credit hours, with a minimum 8 credit hours from the approved list of Upper-Level Robotics Courses and 300-level courses not counted for the Robotics Core requirement
6. Major Design:     ROB 450* Robotics Capstone and
                     TCHNCLCM 350 Technical Communication for Robotics

   * to be developed and approved as part of the Robotics Capstone experience

a. Robotics Major Curricular Graph

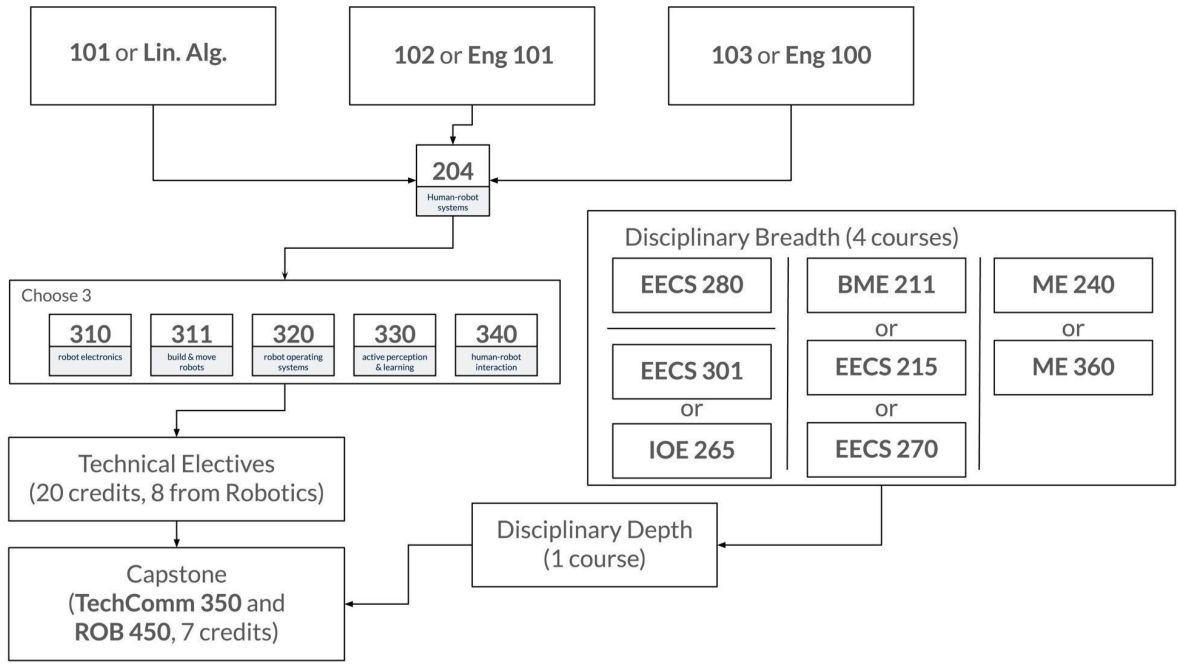

b. Sample Course Schedule for a Major in Robotics

This sample course schedule has in mind a Robotics Major that is oriented towards computing and autonomous robots. Similar sample course schedules can be realized for Robotics Majors oriented towards mechanical systems, electrical systems, aerial robotics, autonomous underwater systems, human factors, project management, and more.

| Sample Undergraduate Robotics Student Schedule | | | Semester # | | | | | | | |
|---|---|---|---|---|---|---|---|---|---|---|
| | Course Name | Credit hours | 1 | 2 | 3 | 4 | 5 | 6 | 7 | 8 |
| Required for all CoE | Robotics 101 or Math 214 | 4 | 4 | | | | | | | |
| | Engineering 100 or Robotics 103 | 4 | 4 | | | | | | | |
| | Engineering 101 or Robotics 102 | 4 | | 4 | | | | | | |
| | Mathematics 115 and 116 | 8 | | 4 | 4 | | | | | |
| | Mathematics 215 or 216 | 4 | | | | 4 | | | | |

| | | | | | | | | | |
|---|---|---|---|---|---|---|---|---|---|
| | Physics 140 and Lab 141 | 5 | | | 5 | | | | |
| | Physics 240 and Lab 241 | 5 | | | | | 5 | | |
| | Chemistry 125/126/130 or 210/211 | 5 | | | | | 5 | | |
| | Intellectual Breadth | 16 | 4 | 4 | 4 | | 4 | | |
| Robotics Program Core | Robotics 204 | 4 | | 4 | | | | | |
| Robotics Intermediate | Robotics 310 | 0 | | | | | | | |
| (choose 3 of 5) | Robotics 311 | 4 | | | | | 4 | | |
| | Robotics 320 | 4 | | | 4 | | | | |
| | Robotics 330 | 4 | | | | 4 | | | |
| | Robotics 340 | 0 | | | | | | | |
| Disciplinary Breadth | EECS 280 - Data Structures | 4 | | | 4 | | | | |
| | IOE 265 - Probability and Stat | 3 | | | | | 3 | | |
| | EECS 215 or 270 or BME 211 | 4 | | | 4 | | | | |
| | ME 240 or 360 | 4 | | | | 4 | | | |
| Disciplinary Depth | NAVARCH 270 | 4 | | | | 4 | | | |
| Capstone | TCHNCLCM 350 | 3 | | | | | | 3 | |
| | Robotics 450* or EECS 467 | 4 | | | | | | | 4 |
| Technical Electives | Upper Level Robotics Electives | 12 | | | | | 4 | 4 | 4 |
| | Flexible Technical Electives | 7 | | | | | 3 | | 4 |
| General Electives | General Electives | 12 | 4 | | | | | 4 | 4 |
| | Total | 128 | 16 | 16 | 16 | 17 | 15 | 16 | 16 | 16 |

*to be developed and approved as part of the Robotics Capstone experience*

# Course Descriptions
## 100 Level Courses

## ROB 101. Computational Linear Algebra
*(3 credits)*
Offered since F20.
Linear algebra and computation as a means for reasoning about data and making discoveries about the world. Topics: The Julia programming language. Systems of linear equations. Vectors, matrices, inverses. Regression. Matrix factorization. Spatial coordinates. Cameras, LiDARS, accelerometers, single-axis gyroscopes, encoders. Optimization and robot perception. What is an ODE.

## ROB 102. Introduction to AI and Programming
*(4 credits)*
Offered since F21.
Algorithms and programming for robotics and artificial intelligence in C++ and high-level scientific programming languages; autonomous navigation and search algorithms; introduction to models of computing through graphs and graph algorithms.

## ROB 103. Robotics Mechanisms
*(4 credits)*
Offered since W21.
Hands-on design, build, and operations of robotic systems. Students, in teams, will build a mobile manipulation robot that can be teleoperated. Students will develop maker-shop skills (3D printing, laser cutting, mill, etc.), programming (C++) and controls, system design and integration, and technical writing. Culmination in friendly competition and final report.

## 200 Level Courses

## ROB 204. Introduction to Human-Robot Systems
*(4 credits)*
Offered since W22.
This foundation in human-robot systems covers identifying and describing how human capabilities and behaviors inform robotic design. We survey theories, methods, and findings from relevant domains (e.g., cognitive/physical ergonomics, psychology, human-centered design), with attention to how these concepts influence robotic systems and design within development teams.

# 300 Level Courses

## ROB 310. Robot Sensors and Signals
*(4 credits)*

Covers practical analog and digital electronics for robotics. Students will: prototype, test, and debug various analog and digital circuits; interface a microcontroller to external circuits; learn to design and prototype circuit boards; interpret data recorded from physical circuits. An exploration of circuits and embedded systems that supports integrated robotic design.

## ROB 311. How to Build Robots and Make Them Move
*(4 credits)*
ROB 311 introduces the fundamentals of mechanical design, control, fabrication, actuation, instrumentation, and computer interfaces required to realize robotic systems. Students will learn to analyze/simulate rigid body kinematics, kinetics, and dynamics, as well as assess the impedance properties of their designs. 'Hands-on' skills will be emphasized in addition to theoretical concepts.

## ROB 320. Robot Operating Systems
*(4 credits)*
Offered since W22.
General computational paradigm for robot operating systems that model, simulate, and control mobile manipulation robots. Composition of full-stack software systems for forward and inverse kinematics, planar path planning, high-dimensional motion planning, maximal coordinate robot simulation, and front-end visualization that work through inter-process communication.

## ROB 330. Localization, Mapping, and Navigation
*(4 credits)*
The development of full-stack autonomous navigation and semantic mapping for mobile robots. Topics include dead reckoning from odometry, sensor modeling of LIDAR and IMUs, simultaneous localization and mapping, semantic scene understanding, and an introduction to deep learning methods for convolutional feature learning and object detection.

## ROB 340. Human-Robot Interaction
*(4 credits)*
Covers psychophysics, modeling a human operator within a control loop, and measuring human performance in the context of robotic systems. These topics support robotic systems in unstructured and unknown environments with a human supporting decision making, mitigating risks and extending capabilities of the human-robot team.

## Technical Communication for Robotics (TCHNCLCM 350)
*(4 credits)*
Teaches students the communication skills needed to support the lifecycle of an engineering project in Robotics, including: communication to understand requirements, articulating those requirements, writing a proposal, communicating with one's team, sharing project updates, and presenting final project work. Includes career documents and best practices for visual communication.

## 400 Level Courses

### ROB 410. Advanced Sensors
*(4 credits)*
ROB 410 covers a multitude of topics in sensing for robotics. The course will cover sensing technologies that allow robots to feel, see, hear, smell, and perceive their environment. This class will also examine sources of noise and errors, passive and active sensing techniques, sensor calibration, and data fusion.

### ROB 411. Robot Controls
*(4 credits)*
A controls course for roboticists by roboticists! This course covers concepts in linear control theory (e.g., stability, controllability), linear optimal control for fully actuated systems, underactuation and model-predictive control (closed-form and computational control), and elementary nonlinear control (e.g., iLQR, feedback linearization).
Disclaimer: This is a unique opportunity to rethink what controls a roboticist needs to know. We look forward to an evolving syllabus that meets the needs of our unique program.

### ROB 412. Neurorobotics
*(4 credits)*
This class is for students interested in controlling robotic devices using neural signals from the body, which is a rapidly emerging area. It's a computational lab class, where students implement a mix of signal processing and machine learning methods on real motor neural data. Topics include: Recording and processing signals from the body; implementing ML algorithms to interpret neural signals and generate motor commands

### ROB 413. Legged Locomotion
*(4 credits)*
Introduction to modeling and control of legged robots. Topics include types of legged robots, models of legged robot kinematics and dynamics, hybrid system models, software tools for developing models, low-dimensional representations of locomotion, optimization tools for gait design, feedback control approaches for agile locomotion, and state estimation in legged robots.

### ROB 415. Multi-Robot Systems
*(4 credits)*
This course covers a variety of topics related to multi-robot systems. Topics include: Graph representation of multi-robot systems; sensing/communication models and hardware; agreement control, formation control via linear methods; connectivity maintenance, collision avoidance via nonlinear methods; safety-critical control using QPs; robustness against system uncertainty; resilience against misinformation.

### ROB 416. Bio-inspired Robotics

*(4 credits)*
The process of how we learn from Nature can be an innovation strategy for translating principles of function, performance and aesthetics from biology to human technology. Diverse teams of students will collaboratively create original bioinspired design projects. Lectures will address the biomimicry design process using case studies across animal diversity.

## ROB 417. Robot Dynamics and Simulation
*(4 credits)*
Modern rigid body dynamics for robot simulation. Analytical methods equation of motion formulation and computational solution techniques applied to mechanical multibody systems. Kinematics of motion generalized coordinates and maximal coordinates. Analytical and computational determination of inertia properties. Collision detection and contact response.

## ROB 420. Mobile Manipulation and Semantic Robotics
*(4 credits)*
Fundamentals of goal-directed autonomous reasoning for robots performing tasks involving human-level dexterity and understanding. Topics include task and motion planning, models of manipulation affordances, language models for robot understanding, and methods for learning from experience, demonstration, and human guidance. Project-driven course with significant programming and analytical methods.

## ROB 421. Optimal Robotics
*(4 credits)*
Introduction to optimization models and their applications to robotics problems ranging from control to planning to perception to learning. The emphasis will be on understanding which problems are numerically tractable and how to solve them. Assignments will focus on implementing and solving robotics relevant optimization problems.

## ROB 422. Introduction to Algorithmic Robotics
*(4 credits)*
An introduction to the algorithms that form the foundation of robot planning, state estimation, and control. Topics include optimization, motion planning, representations of uncertainty, Kalman and particle filters, and point cloud processing. Assignments focus on programming a robot to perform tasks in simulation.

## ROB 423. Autonomous Vehicles
*(4 credits)*
The objective of this course is to provide students with an introduction to the basic algorithms used for control (lane keeping, ACC, trajectory design via convex and nonlinear programming methods) and perception (SLAM, deep learning, computer vision) that are typically used in self-driving cars.

## ROB 430. Probabilistic Robotics
*(4 credits)*
Theory and application of probabilistic techniques for autonomous mobile robotics. This course will present and critically examine contemporary algorithms for robot perception and navigation, including Bayesian filtering; stochastic representations of the environment; motion and sensor models for mobile robots; algorithms for mapping, localization; application to autonomous marine, ground, and air vehicles. Individual assignments (coding + theory) and Final Group Project.

## ROB 431. Robot Learning
*(4 credits)*
This course covers concepts at the intersection of machine learning, dynamics, and control with emphasis on robotics applications. Topics covered include representation learning for dynamics and controls (e.g., latent and explicit representations), domain adaptation and sim2real transfer, handling model uncertainty, policy/controller learning, and robot learning design. Assignments will include theory and implementation of state-of-the-art algorithms using simulators and real-world data.

## ROB 432. 3D Robot Perception
*(4 credits)*
This course covers advanced topics in autonomous perception, navigation and mapping for mobile robots. Topics include stereo cameras, multi-view geometry, 3D mapping and reconstruction, semantic simultaneous localization and mapping and deep learning methods for 3D point cloud processing. Assignments will focus on implementation of state-of-the-art algorithms with testing on real data.

## ROB 440. Human-Robot Collab
*(4 credits)*
The goal of this course will be to design successful collaborations between humans and robots. Collaboration will be broadly defined to include settings such as homes, hospitals, offices, space exploration and manufacturing. The course will employ an interdisciplinary perspective comprising classical robotics, human-computer interaction, design, and social and cognitive psychology.

## ROB 441. Physical HRI
*(4 credits)*
ROB 441 will introduce the modeling and analysis tools relevant to the design and deployment of robots that make physical contact with humans for the purposes of collaboration, shared control, teleoperation, and two-way haptic communication. Topics in biomechanics and interaction dynamics will be introduced to guarantee stability, safety, and performance as targeted by design.

## ROB 442. Ethics AI & Robotics
*(4 credits)*
The course examines ethics from several perspectives. Trust: cooperation, society, and corporate entities. Safety: Autonomous vehicles and their decisions. Surveillance and privacy; Bias and fairness; Existential risk. Concepts are drawn from philosophical ethics, engineering design, law, economics, evolution, history, and human development. How can knowledge relevant to making ethical decisions be represented computationally in a knowledge base, and how can it be acquired?

## ROB 450. Robotics Capstone
*(4 credits)*
This course brings together all of the concepts, enabling tools, and state-of-the-art resources within a team-based project class. Students will work on industry or faculty mentored projects to solve important societal challenges, meet critical customer needs, or create innovative robotic systems that inspire and provide a learning platform for the next generation of roboticists.

## ROB 464. Hands-on Robotics
*(4 credits)*
A hands-on, project-based introduction to the principles of robotics and robot design. Multiple team projects consisting of design and implementation of a robot. Theory: motors, kinematics & mechanisms, sensing/filtering, planning, pinhole cameras. Practice: servo control, project management; fabrication; software design for robotics. Significant after-hours lab time investment.

## ROB 470. Experimental UAS
*(4 credits)*
This course will have a similar structure to ROB 550 (Robot Systems Laboratory) but focuses on Unmanned Aircraft systems (UAS). There are two lab projects:
Project 1: Quadrotor Lab (Quadlab)
Students model propulsion/mass properties and calibrate IMU sensors; students perform a series of experiments to achieve autonomous flight
Project II: Hybrid Plane Lab (Planelab)
Students design, build, and test a custom small UAS with fixed-wing lift and vertical thrust
Lab, wind tunnel and M-Air tethered flight tests

## ROB 471. Marine Robotics
*(4 credits)*
Overview of marine robotic systems, including autonomous surface vehicles, remotely operated vehicles, and autonomous underwater vehicles. Topics include vehicle design, kinematic and dynamic modeling, control, sensing, and navigation. Examples draw from real robotic missions across a range of applications from inspection of critical subsea infrastructure to exploration of ocean worlds

ROB 472. Space Robotics
*(4 credits)*
The class is an in-depth exposition of space robotics; it discusses autonomous spacecraft, planetary and lunar rovers, path planning, robotic arms and robonauts. The class first considers autonomous vehicles; current launch vehicles are largely autonomous, and are studied first. Missions such as automatic rendezvous, docking and undocking, inspection, repair and refueling are considered.

Planetary rovers are considered next, including the design of suspension systems and robotic arms, approach and landing protocols, rover navigation, and path planning strategies (occupancy grids, A* planning, probabilistic roadmaps, RRTs, interpolation, potential fields and SLAM). Articulated robot arms are discussed, including examples from the ISS and the Curiosity rover. Manipulator configurations, redundancy, joints and workspaces are considered. The NASA/GM Robonaut R2 is treated as an example. Forward and inverse kinematics transformations between joints, tool position and tool orientation for robot arms are discussed, and control strategies considered in detail.

Each offering, an existing mission is used as an example scenario, including mission design, configuration, autonomous navigation and targeting, flight control systems, and mission specific algorithms.

# Planned Course Offerings and Instruction Needs

As shown below, the proposed degree is an in-person on-campus offering to begin for Fall Semester 2022. We plan to increase course offerings and achieve a steady state in Fall 2024. The number of instructors will similarly be ramped up, starting from 12 instructors in Fall 2022.

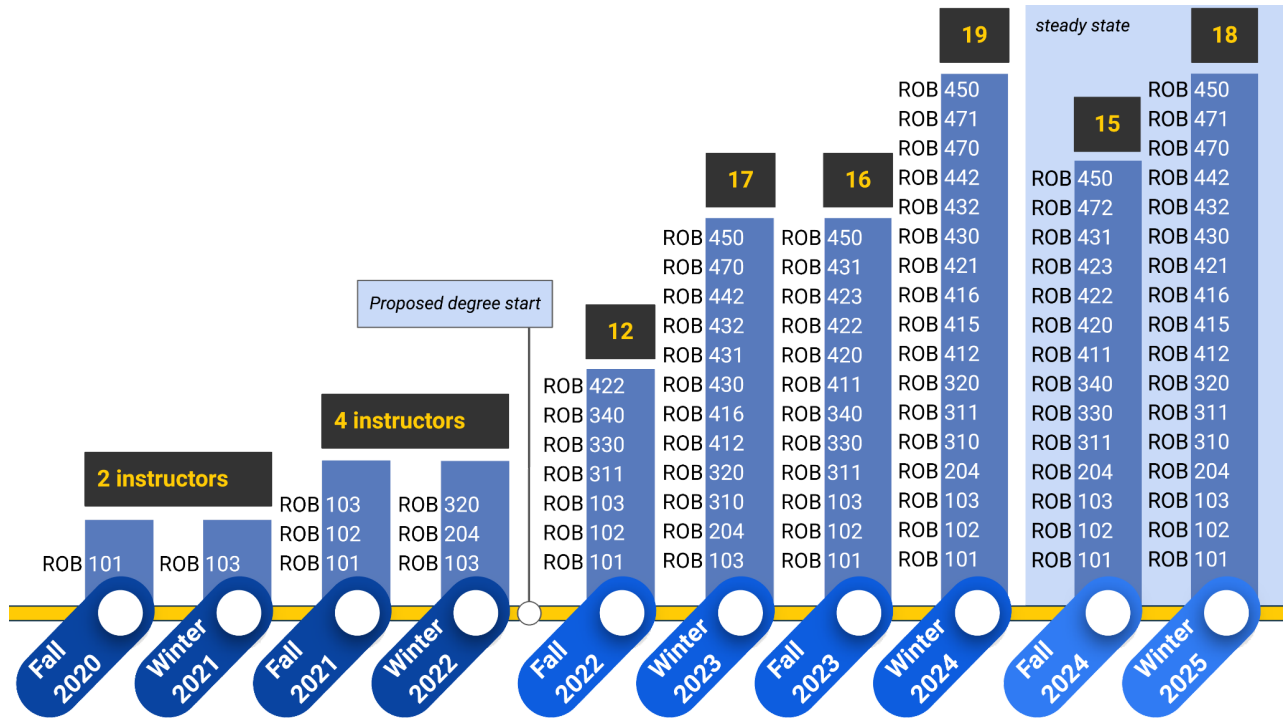

# Statement on Faculty Qualifications

Our robotics curriculum will be taught by world-class faculty doing cutting edge research in robotics, building on the existing foundation of The U-M Robotics Institute. The Robotics Institute currently comprises an interdisciplinary group of 30 core faculty that span 12 departments, coupled with 42 affiliate faculty that contribute to the breadth and excellence of robotics research. Robotics faculty research expenditures continue to show significant growth, outpacing the addition of new faculty. Indeed, over the past five years, research expenditures have increased from $5.6M to $16.0M, nearly tripling despite the addition of only five faculty members (~20% new faculty). The Institute emphasizes 'Full Spectrum Robotics,' with a broad array of faculty research interests, including human-robot interaction, legged and rehab robotics, artificial intelligence, autonomous and connected vehicles, dexterous manipulation, among many other areas.

The Institute offers a Graduate Program in Robotics with both MS and PhD degrees. Graduate students first matriculated in Fall 2014, and the program currently includes 231 students (157 MS and 74 PhD). Since its inception, applications have risen to over 1,000 in 2020.

When the Robotics Department is launched, there will be between 12 and 15 College of Engineering tenure or tenure-track faculty who will transfer tenure to the new Robotics Department, joining the existing two lecturers in Robotics. The existing departments are insisting that the faculty transition in a phased manner. In addition, the College will provide hiring slots.

The table below shows the planned transitions of tenure slots and the hiring of new faculty. We may need to call on transferring faculty to teach Robotics undergraduate courses before they transfer tenure and all of their research dollars to Robotics. This is under negotiation with the Office of the Associate Dean for Academic Affairs (ADAA).

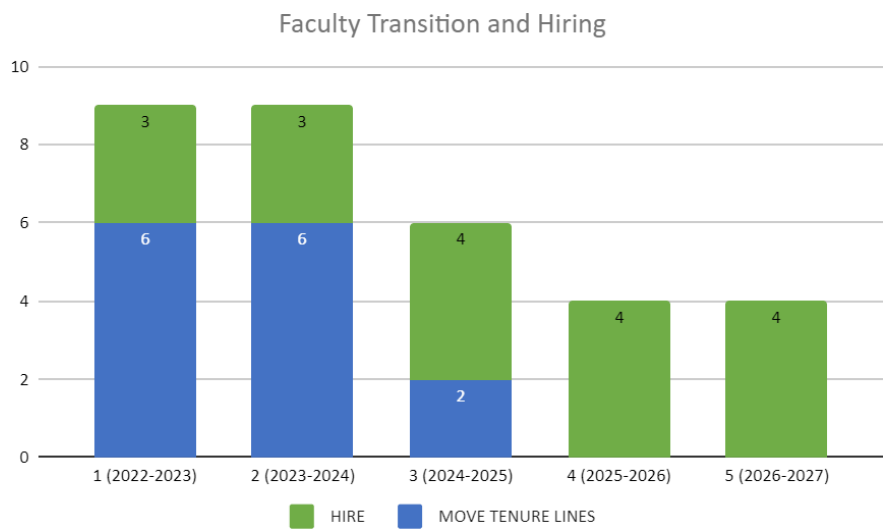

# Faculty Bios

The following faculty will teach Robotics courses as a part of the Robotics undergraduate degree program.

| Faculty | Expertise |
|---|---|
| Abdon Pena-Francesch | biomaterials science, soft matter physics, and nanotechnology |
| Anouck Girard | dynamics, control, and optimization of autonomous vehicles |
| Brent Gillespie | haptic interface and robotics |
| Chad Jenkins | computational reasoning and perception |
| Chandramouli Krishnan | rehabilitations to mitigate neuromuscular impairments |
| Cynthia Chestek | brain machine interface systems |
| Dawn Tilbury | control theory and applications, such as manufacturing systems |
| Deanna Gates | rehabilitation and biomechanics |
| Dimitra Panagou | nonlinear systems, control and estimation; multi-agent systems |
| Dmitry Berenson | learning and motion planning for manipulation |
| Ella Atkins | autonomous aerospace systems |
| Elliott Rouse | development of exoskeletons and robotic prostheses |
| Jason Corso | high-level computer vision |
| Jessy Grizzle | control of bipedal robots |
| Katherine Skinner | autonomy in dynamic, unstructured, or remote environments |
| Kira Barton | control theory, manufacturing, nano-scale printing |
| Leia Stirling | human-robot interaction |
| Lionel Robert | human-robot teams |
| Maani Ghaffari | robust autonomous robotic exploration |
| Nadine Sarter | cognitive ergonomics |
| Necmiye Ozay | dynamical systems, control, optimization and validation |
| Nima Fazeli | robotic manipulation and embodied intelligence |
| Peter Gaskell | robotics platforms for engineering education |
| Ram Vasudevan | optimization and control of nonlinear and hybrid dynamical systems |
| Robert Gregg | control mechanisms of bipedal locomotion |
| Talia Moore | biomechanics and bioinspiration |

# Admissions Requirements and Projected Enrollment

**Enrollment**

We estimate undergraduate enrollment will be 435 at the lower end, according to market analysis and surveys. As many as 40 percent of current students who responded to the survey expressed interest in a U-M Robotics major or minor, while 25 percent of respondents who enrolled elsewhere indicated that they might have decided differently if U-M currently offered these degree programs.[5]

Based on a conservative interpretation of these results, we estimate that 10 percent of incoming College of Engineering freshmen will declare Robotics. Assuming no attrition, no incoming transfers, and 4-year graduation rates, the department will enroll 435 undergraduate students in steady state. Of course, this enrollment level needs to be achieved over time. Below, we show the proposed enrollment in the initial five-year ramp-up period.

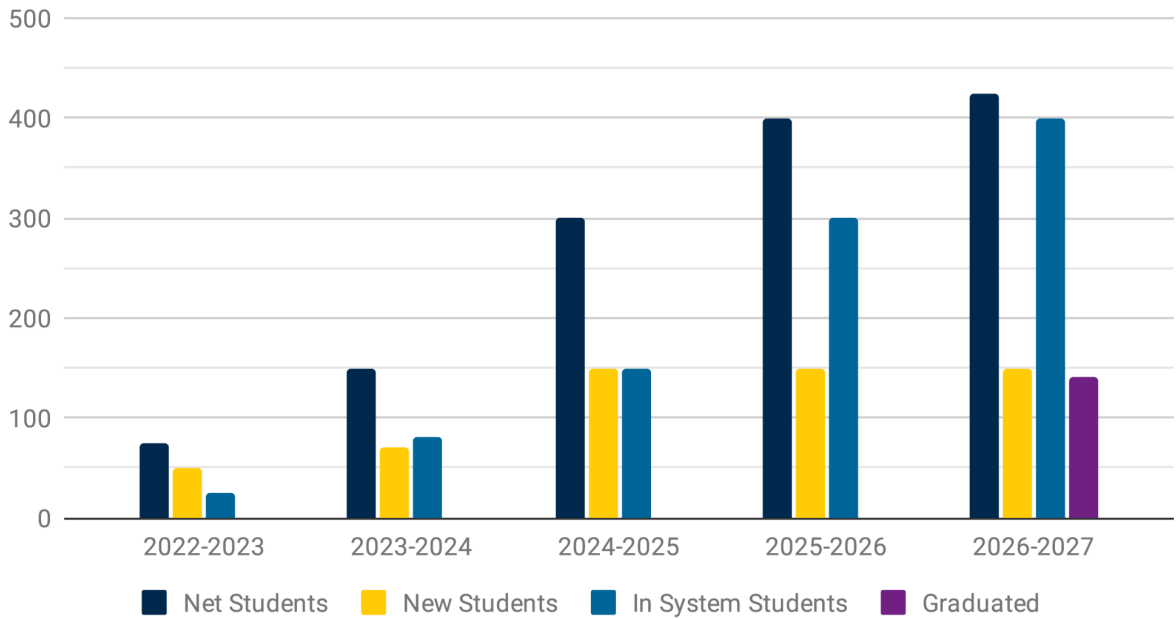

*The proposed ramp-up of undergraduate student enrollment. Net students is the sum of new and in-system students less the number of graduated students.*

---

[5] Nexus report, pg. 26

# Accreditation & Program Assessment

## Accreditation

The Robotics Undergraduate Program is planning to take a leading role in defining the standards for robotics as it takes shape into academic programs across the country and the world. We believe robotics is rapidly evolving as a nascent discipline, where accreditation could inhibit growth and cause stagnation. Accreditation by existing bodies, such as the Accreditation Board for Engineering and Technology, Inc., are well suited for programs in robotics engineering or mechatronics that are more geared to be extensions of traditional engineering disciplines that have been well established over decades or centuries. Although we are not planning to seek undergraduate accreditation at the present time, U-M Robotics is dedicated to continually improving its programs through regular feedback from students and experts in both academia and industry. The following sections outline the proposed measures through which the Robotics Undergraduate Program will be assessed and evaluated.

## Six-year Undergraduate Program Review

In concert with the planned Robotics Department, the Robotics Undergraduate Program is proposed to be reviewed every six years. This program review will consist of an internal review and an external review, each of which will produce evaluation reports that are submitted to the College of Engineering. The internal review will be done by a committee of selected faculty to summarize the current state of the Robotics Undergraduate Program and evaluate its strengths, weaknesses, and opportunities. This internal review will be followed by an external review performed by esteemed individuals in the field of robotics. It is expected that at least one member of this committee will also be a member of our standing Diversity Advisory Board. The report generated from this external review will be submitted to the College of Engineering.

## Diversity Advisory Board

We plan to have a group of highly distinguished and accomplished individuals from academe, industry, and government that will meet with us annually. Their expertise will help us in constructing and evolving a curriculum that is inclusive, fair, technically sound, meets the needs of professional practice as well as leading graduate programs, and could potentially be adopted by other departments and universities. We already have initial commitments from several members, and also plan to include the College of Engineering Associate Dean of Undergraduate Education to provide advice for socializing our ideas at the 100-level with all of engineering at U-M.

## Longitudinal Student Surveys

The Robotics Undergraduate Program firmly believes that wise student feedback comes longitudinally with time and perspective. As such, we will conduct annual surveys of the group of individuals that have enrolled in a U-M Robotics course at any period of time. In addition to current students and dedicated roboticists, this group surveyed includes alumni, students that may have considered robotics but took a different path, and engineering students whose studies overlap with Robotics. These surveys will be designed to gather information about student satisfaction, ideas for potential improvements and innovations, and program effectiveness in the professional outcomes longitudinally. These longitudinal surveys are expected to provide a foundation for continued formulation of our program goals and reporting about our effectiveness towards those goals, similar in spirit to the [U-M CSE Climate, Diversity, Equity, and Inclusion Annual Reports](#).

## Requirements for Funding, Space, and Equipment

Through the proposal planning and implementation processes it has been confirmed that funding is available to provide adequate support for the proposed new program. Resources allocated to the new initiative will not have a negative impact on existing programs. The recommendations made represent an effective and efficient use of institutional resources, which will be part of the ongoing infrastructure of the College of Engineering. Space needs, including specialized instructional lab spaces and equipment were addressed in the construction of a new building (please see below). No additional major expenditures for computers, laboratory space, equipment or library holdings is required. A planning committee made recommendations to the U-M College of Engineering for hiring additional faculty (please see pg. 30) and administrative staff, and incorporating the program into the existing funding infrastructure. At this time Associate Dean Millinchick is responsible for implementing the recommendations related to the undergraduate degree, Associate Dean Ceccio is responsible for implementing recommendations related to faculty transition and hiring, and Executive Director Mero is responsible for implementing recommendations related to funding and administration.

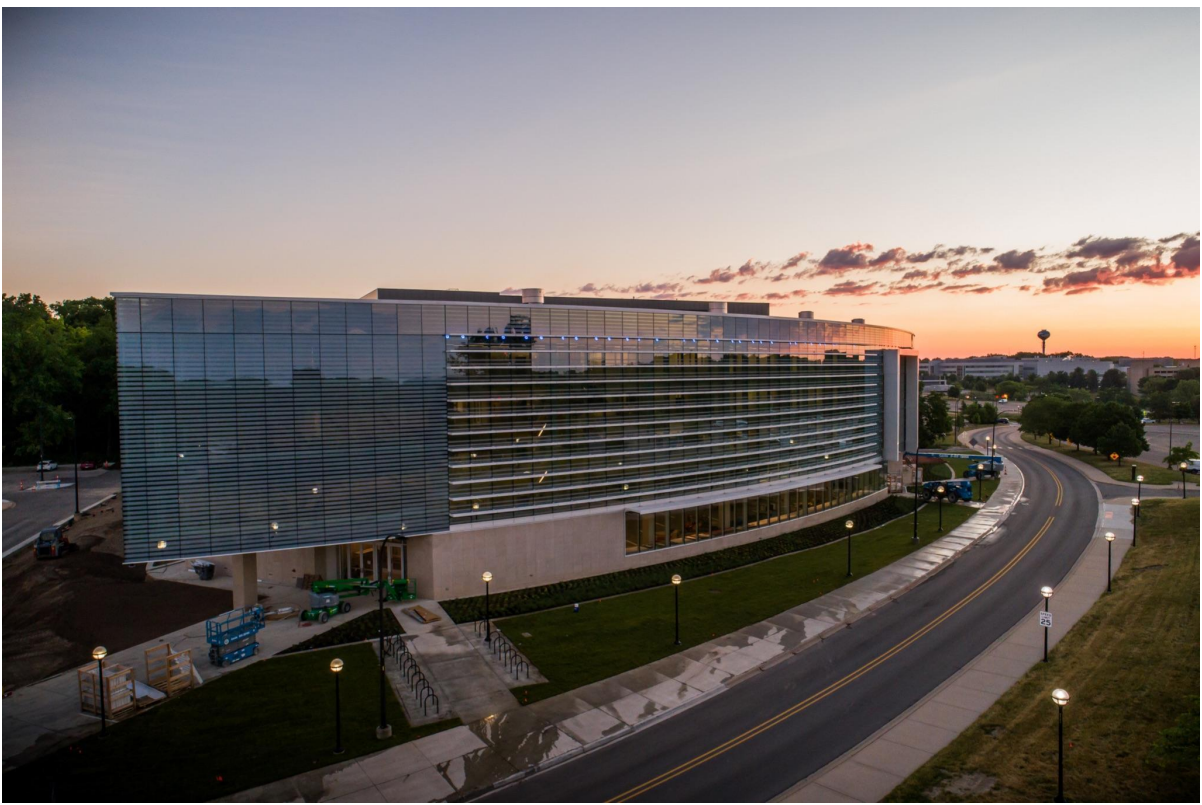

**Funding and Equipment**
Funding for the Robotics undergraduate program will be part of the annual budget for the U-M Robotics Department. At steady-state, the Robotics Department will participate in the College of Engineering's regular funding model for all departments. A team is currently developing a transitional funding plan for Robotics as both the department and undergrad program ramp up activity. We will use the College of Engineering's annual budget request process to request funding for special projects, new strategic initiatives, additional staffing, and other costs as they arise. In addition we will collaborate with the College of Engineering to secure gift funding, which can help supplement special projects and initiatives.

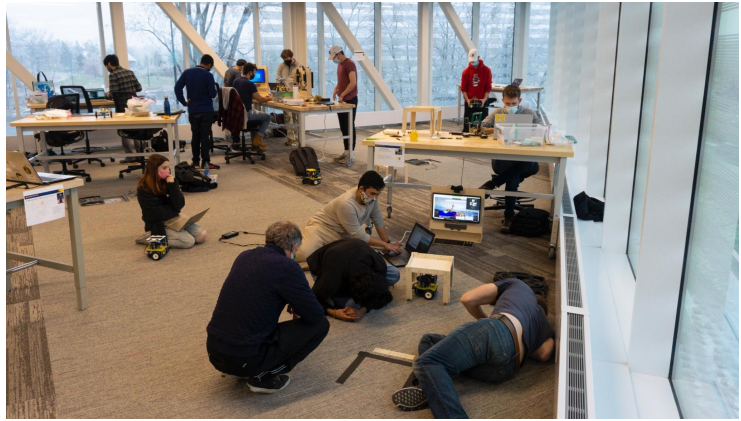

Two specific components of the robotics undergraduate major that will require funding are course support staff and course supplies. The course support staff include Graduate Student Instructors (GSIs) and Instructional Aides (IAs). Our goal is to provide course support staff for the majority of the undergraduate courses included in the new major. Course support staff make courses more accessible by helping students better engage with the material and provide further in-depth and small-group or individual instruction.

The curriculum includes many lab courses that include a number of hands-on, building, and fabricating components. These courses will require consumable materials, and most of the required supplies are inexpensive. We will purchase course supplies based on expected enrollment numbers in order to adequately run each lab course. As the undergraduate major ramps up, we will track spending on course supplies in order to include a sufficient amount in the program's annual budget.

**Space**
The Ford Motor Company Robotics Building is a four-story, $75 million, 134,000-square-foot complex situated on North Campus. Its first three floors hold custom U-M research labs for robots that fly, walk, roll and augment the human body—as well as classrooms, offices and makerspaces. Through a unique agreement, the fourth floor houses Ford's first robotics and mobility research lab on a university campus, as well as 100 Ford researchers and engineers.

While the Ford Motor Company Robotics Building will be the main home for the robotics program, undergraduates will have use of the entire U-M central and north campuses with similar world-class research and education buildings.

Building features includes:

- The Ronald D. and Regina C. McNeil Walking Robotics Laboratory for developing and testing legged robots, with an in-ground treadmill that can hit 31 mph and a 20% grade, as well as carry obstacles. Walking robots could aid in disaster relief and lead to better prosthetics and exoskeletons. 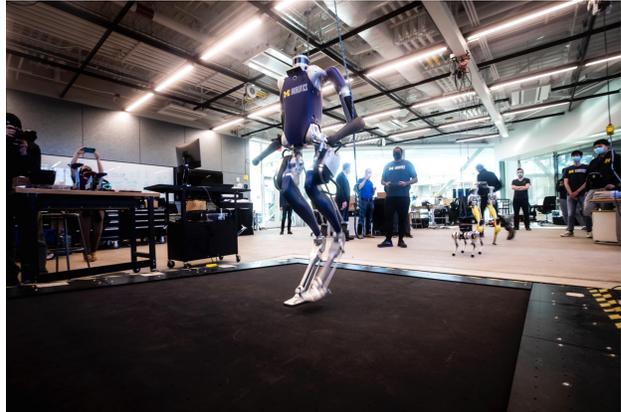
- A Rehabilitation Lab for advanced prosthetics and robotic controls, with a movable "earthquake platform" that can tilt in any direction, while force-feedback plates measure ground contact.
- A three-story fly zone to test drones and other autonomous aerial vehicles indoors, before moving to the adjacent outdoor M-Air research facility. Autonomous aerial vehicles could perform safer inspection of infrastructure like windmills and bridges.
- A Mars yard, designed with input from planetary scientists at U-M and NASA, to enable researchers and student teams to test rover and lander concepts on a landscape that mimics the Martian surface.
- An AI-designed "robot playground" outdoor obstacle course for testing robots on stairs, rocks, and water, surrounded by motion capture cameras. 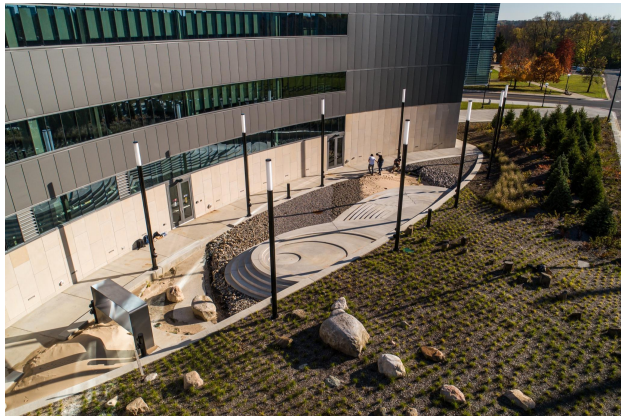
- High-bay garage space for self-driving cars, located just down the road from the [Mcity Test Facility](#), for putting connected and automated vehicles through the paces in simulated urban and suburban environments.

Lab space includes:
- 23,647 sq ft
- 19 assigned PI's
- Support for an additional 9 PI's

There are also two instructional labs (FMCRB 2010 and 2020) with a capacity of 32 students each. In the Maker Space, the Student Team area is being configured to support 24 students per lab section.

**Library**

The library and database needs of this new major are covered by existing U-M Library resources in and beyond the Art, Architecture & Engineering Library (AAEL). In addition, a dedicated librarian in the AAEL has been assigned to support the needs of the Robotics program.

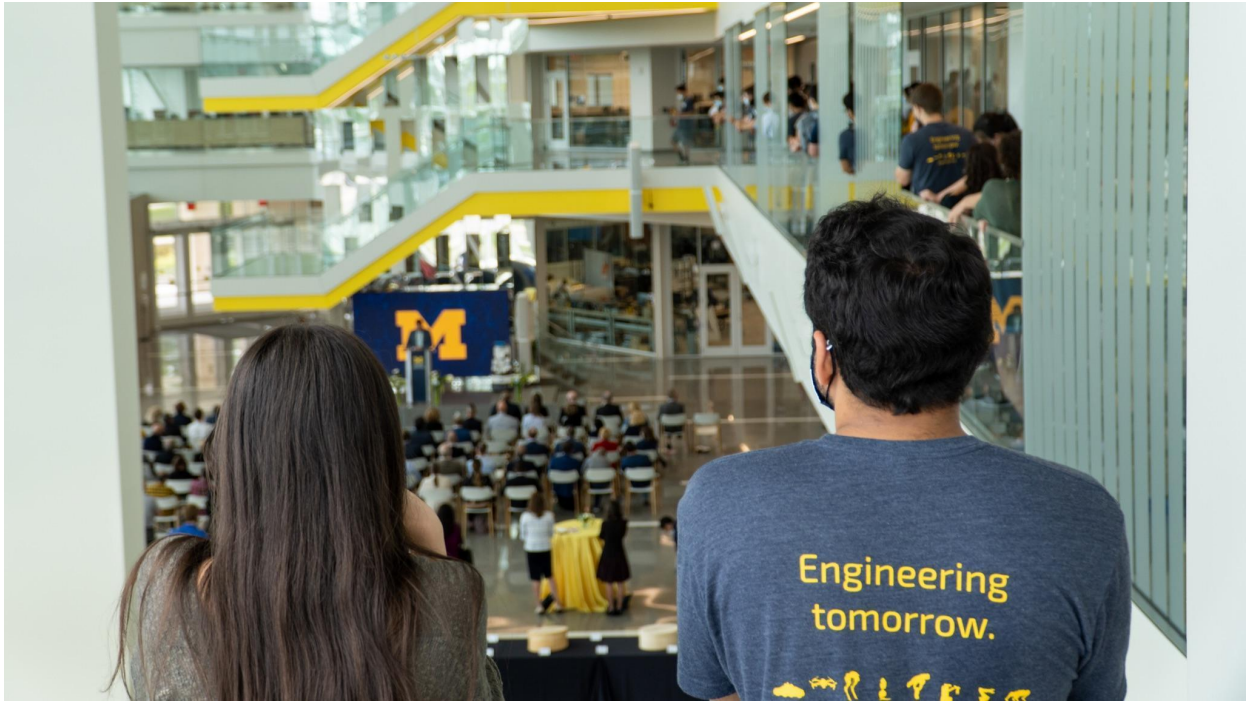